\documentclass[final,5p,times,twocolumn]{elsarticle}

\usepackage{tikz}
\usepackage{graphicx}
\usepackage{pstricks}
\usepackage{verbatim}
\usepackage{psfrag}
\usepackage[english]{babel}
\usepackage{amsmath}
\usepackage{amssymb}
\usepackage[font=small]{caption}
\usepackage{fancybox}
\usepackage{steinmetz}
\usepackage{subcaption}
\usepackage{bodegraph}
\usepackage{gensymb}
\usepackage{algorithm,algorithmic}
\usepackage{amsfonts}
\usepackage{dsfont}
\usepackage{siunitx}
\usepackage{lscape}
\usepackage{enumerate}
\usepackage{lmodern}
\usepackage[T1]{fontenc}
\usepackage[framed]{matlab-prettifier}
\usepackage{mathtools}
\usepackage{color} 
\usepackage{xcolor,colortbl}
\usepackage{colortbl}
\usepackage{multirow, hhline}
\usepackage{url}
\usepackage{hyperref}
\usepackage{multicol}
\usepackage{supertabular}
\usepackage{balance}
\usepackage{lineno,pifont}
\usepackage{multirow}
\usepackage{xcolor}
\usepackage{xcolor}

\definecolor{miored}{rgb}{1.0, 0.0, 0.0}

\def\cN{{\cal N}}

\newcommand{\TesiPubblica}{1=0}           
\newcommand{\ParteSensibile}[2]{\ifnum\TesiPubblica\relax#2\else#1\fi}

\graphicspath{{Images/}}

\newcommand{\MAX}{{\mbox{\tiny MAX}}}

\newcommand{\yorig}{\y(t)}
\newcommand{\dyorig}{\dot{\y}(t)}
\newcommand{\ddyorig}{\ddot{\y}(t)}

\newcommand{\sfk}{\s_{k}}

\newcommand{\yr}{\y_r}

\newcommand{\yl}{\y_{L}}
\newcommand{\ylscal}{y_{L}}

\newcommand{\tr}{\t_T}

\newcommand{\yf}{\y_\Delta}
\newcommand{\yfk}{\y_{\Delta,k}}

\newcommand{\Yfk}{\Y_{\Delta,k}}

\newcommand{\dy}{\dot{\y}}
\newcommand{\ddy}{\ddot{\y}}
\newcommand{\yp}{\y^\star}
\newcommand{\dyp}{\dot{\y}^\star}
\newcommand{\ddyp}{\ddot{\y}^\star}

\newcommand{\yt}{\widetilde{\boldsymbol{y}}}
\newcommand{\dyt}{ \dot{\widetilde{\boldsymbol{y}}} }
\newcommand{\ddyt}{ \ddot{\widetilde{\boldsymbol{y}}} }

\newcommand{\qp}{\q^\star}
\newcommand{\dqp}{\dot{\q}^\star}
\newcommand{\ddqp}{\ddot{\q}^\star}

\newcommand{\yrscal}{y_r}

\newcommand{\tfscaluno}{t_{\Delta_0}}
\newcommand{\tfscaldue}{t_{\Delta_1}}
\newcommand{\tfscaltre}{t_{\Delta_2}}
\newcommand{\tfscalquattro}{t_{\Delta_3}}
\newcommand{\tfscalcinque}{t_{\Delta_4}}
\newcommand{\tfscalsei}{t_{\Delta_5}}
\newcommand{\tfscalsette}{t_{\Delta_6}}
\newcommand{\tfscalotto}{t_{\Delta_7}}
\newcommand{\tfscalnove}{t_{\Delta_8}}

 \newcommand{\tfscalk}{t_{\Delta_{k}}}

\newcommand{\trscaluno}{t_{T_0}}
\newcommand{\trscaldue}{t_{T_1}}
\newcommand{\trscaltre}{t_{T_2}}
\newcommand{\trscalquattro}{t_{T_3}}
\newcommand{\trscalcinque}{t_{T_4}}
\newcommand{\trscalsei}{t_{T_5}}
\newcommand{\trscalsette}{t_{T_6}}
\newcommand{\trscalotto}{t_{T_7}}
\newcommand{\trscalnove}{t_{T_8}}
\newcommand{\trscaldieci}{t_{T_{9}}}
\newcommand{\trscalundici}{t_{T_{10}}}
\newcommand{\trscaldodici}{t_{T_{11}}}
\newcommand{\trscaltredici}{t_{T_{12}}}
\newcommand{\trscalquattordici}{t_{T_{13}}}
\newcommand{\trscalquindici}{t_{T_{14}}}

\definecolor{Gray}{gray}{0.9}
\definecolor{verdedd}{rgb}{0.0, 0.5, 0.0}
\newtheorem{Definizione}{\bf Definition}
\newtheorem{Prop}{\bf Property}
\newcommand{\mat}[2]{\left[\begin{array}{#1} #2 \end{array}\right]}

\definecolor{myorangenew}{rgb}{1 0.49 0}
\definecolor{mygreennew}{rgb}{0.31 0.78 0.47}
\definecolor{mygreynew}{rgb}{0.6 0.6 0.6}
\definecolor{bluD}{rgb}{0.09, 0.45, 0.81}
\definecolor{bluD_l}{rgb}{0.29, 0.65, 1}
\definecolor{nero}{rgb}{0, 0, 0}
\definecolor{bianco}{rgb}{1, 1, 1}

\definecolor{bluG}{cmyk}{100 85 0 0}
\definecolor{verdeG}{cmyk}{75 0 100 0}

\def\e{{\boldsymbol e}} \def\f{{\boldsymbol f}} \def\g{{\boldsymbol g}}

\def\q{{\boldsymbol q}}  \def\s{{\boldsymbol s}} \def\t{{\boldsymbol t}}
\def\u{{\boldsymbol u}}
 \def\x{{\boldsymbol x}}
\def\y{{\boldsymbol y}} \def\z{{\boldsymbol z}}

\def\A{{\boldsymbol A}} \def\B{{\boldsymbol B}}  
\def\E{{\boldsymbol E}} \def\F{{\boldsymbol F}} \def\G{{\boldsymbol G}} 
\def\I{{\boldsymbol I}}   
   
 \def\R{{\boldsymbol R}}  
   
\def\Y{{\boldsymbol Y}} 

\newtheorem{Remar}{\bf Remark}

\newcommand{\gcheck}{{\color{green}\ding{51}}}
\newcommand{\rcross}{{\color{red}\ding{55}}}

\DeclareMathOperator*{\argmin}{argmin}
\DeclareMathSymbol{\boldmu}{\mathord}{letters}{15}

\journal{Robotics and Autonomous Systems}

\begin{document}

\begin{frontmatter}

\title{{\black Phase-Independent Dynamic Movement Primitives With Applications to Human-Robot Co-manipulation and Time Optimal Planning}}

\author{Giovanni Braglia\corref{mycorrespondingauthor}}
\cortext[mycorrespondingauthor]{Corresponding author}
\ead{giovanni.braglia@unimore.it}

\author{Davide Tebaldi}
\ead{davide.tebaldi@unimore.it}

\author{Luigi Biagiotti}
\ead{luigi.biagiotti@unimore.it}

\affiliation{organization={University of Modena and Reggio Emilia, Department of Engineering Enzo Ferrari},
            addressline={Via Pietro Vivarelli 10},
            city={Modena},
            postcode={41125},
            country={Italy}}

\begin{abstract}

{\black Dynamic Movement Primitives (DMP) are an established and efficient method for encoding robotic tasks that require adaptation based on reference motions. Typically, the nominal trajectory is obtained through Programming by Demonstration (PbD), where the robot learns a task via kinesthetic guidance and reproduces it in terms of both geometric path and timing law. Modifying the duration of the execution in standard DMPs is achieved by adjusting a time constant in the model.\\ This paper introduces a novel approach to fully decouple the geometric information of a task from its temporal information using an algorithm called spatial sampling, which allows parameterizing the demonstrated curve by its arc-length. This leads to the definition of the Geometric DMP (GDMP). The proposed spatial sampling algorithm guarantees the regularity of the demonstrated curve and ensures a consistent projection of the human force throughout the task in a human-in-the-loop scenario. GDMP exhibits phase independence, as its phase variable is no longer constrained to the demonstration's timing law, enabling a wide range of applications, including phase optimization problems and human-in-the-loop applications. Firstly, a minimum task duration optimization problem subject to velocity and acceleration constraints is formulated. The decoupling of path and speed in GDMP allows to achieve optimal time duration without violating the constraints. Secondly, GDMP is validated in a human-in-the-loop application, providing a theoretical passivity analysis and an experimental stability evaluation in co-manipulation tasks. Finally, GDMP is compared with other DMP architectures available in the literature, both for the phase optimization problem and experimentally with reference to an insertion task, 
showcasing the enhanced performance of GDMP with respect to other solutions. 
}

\end{abstract}

%


\end{frontmatter}

\section{Introduction}
\label{sec1}


{\black In recent years, the interest in collaborative robotics has been significantly rising, as robots are starting to work with humans rather than performing separate tasks~\cite{zhang2020development}. 
This scenario is typically referred to as Human-Robot Interaction (HRI), and a lot of research interest has been directed towards the development of novel algorithms and techniques to enhance the human-robot communication, coexistence and collaboration in various settings~\cite{ashish,6,30}.
An important aspect of HRI applications is represented by their accessibility to 
non-expert users~\cite{calinon2007learning,33}, where techniques such as Imitation Learning~\cite{schaal1999imitation},
or Programming by Demonstration (PbD)~\cite{33}, come into play.
These methods involve a demonstration phase and rely on encoding, planning and control strategies to correctly reproduce the desired task~\cite{hovland1996skill,35}.}
One of the most widespread encoding/planning techniques available in the literature are Dynamic Movement Primitives (DMP)~\cite{4}, which were first introduced in~\cite{5} as a way to encode motor
movement with convenient stability properties.
DMP consist of a second-order system with the addition of a forcing term~\cite{8} and can be classified into rhythmic or discrete DMP, depending on whether the attractor system is a limit cycle or a point~\cite{4}, respectively.
Discrete DMP are typically formulated as follows~\cite{4,8}:
\begin{align}
  &  \tau \dot{z}(t) = \alpha [ \beta(g-y(t))- z(t) ] + f(s(t)), \label{ts1}
\\[1mm]
 &   \tau \dot{y}(t) = z(t),
\label{ts2}
\\[1mm]
  &  \tau \dot{s}(t) = - \delta s(t),
\label{cs}
\end{align}
where Eqs.~\eqref{ts1}-\eqref{ts2} are referred to as the Transformation System (TS), whereas~\eqref{cs} is referred to as the Canonical System (CS). The parameter $\tau$ is a time constant, $\alpha$ and $\beta$ are system parameters 
chosen to have a critically damped  TS (i.e. $\beta = \alpha/4$), 
$g$ is the target goal and the forcing term $f(s(t))$ shapes the generated trajectory $y(t)$ through the TS dynamics. The TS can be rewritten as
\begin{equation}
\tau^2\ddot y(t) + \alpha \tau \dot y(t) + \alpha \beta(y(t)-g) = f(s(t)),
\label{eq:SecondOrderSystem}
\end{equation}
representing a globally stable second-order system characterized by the unique equilibrium point $(y(t),\dot{y}(t))=(g,0)$ when $f(s(t))=0$.
The CS \eqref{cs} is a first-order system representing the evolution of the phase variable $s(t)$, with $\delta$ being a positive constant~\cite{8}.
The forcing term $f(s(t))$ in \eqref{ts1} is typically expressed as the linear combination of radial basis functions~\cite{5}:
\begin{equation}
\label{fs}
    f(s(t)) = \sum_{i=1}^{N} \omega_i \phi_i(s(t)) \cdot s(t)\cdot(g-y_0),    
\end{equation}
where
\begin{equation} \label{rbf}
\phi_i(s(t)) \!=\! \frac{b_i(s(t))}{\sum_{j=1}^{N} b_j(s(t))}, \,\, b_i(s(t)) \!= \!\textrm{exp} \!\left(\! \frac{-\!(s(t)\!-\!c_i)^2}{2h}\!\right).
\end{equation}
Note that, since $s(t)\rightarrow 0$ when $t\rightarrow \infty$ from \eqref{cs}, $f(s(t)) \rightarrow 0$ from \eqref{fs}, therefore its impact in the TS \eqref{ts1}-\eqref{ts2} is limited in time. The constant $(g-y_0)$ in \eqref{fs} is a modulation term enabling the TS \eqref{ts1}-\eqref{ts2} to be an orientation-preserving homeomorphism with respect to changes in the initial and goal positions $y_0$ and $g$~\cite{5,73}. The weight coefficients $\omega_i$ in \eqref{fs} are calculated by Locally Weighted Regression (LWR)~\cite{atkeson1989using} as follows: 
\begin{equation}
\label{lwr}
    \omega_i = \argmin_{\omega_i} \sum_{t=0}^{T_f} \phi_i(s(t))[f_r(t) - \omega_i\xi(t)] ^2,
\end{equation}
where $\xi(t) = s(t)(g-y_0)$ and the nominal forcing term $ f_r(t)$ is computed from \eqref{eq:SecondOrderSystem} by using the reference position, velocity, and acceleration trajectories $y_r(t), \dot{y}_r(t)$ and $ \ddot{y}_r(t)$:
\begin{equation}
\label{fd}
f_r(t) = \tau^2 \ddot{y}_r(t) + \alpha \tau \dot{y}_r(t) + \alpha \beta(y_r(t)-g_r),
\end{equation}
where $g_r = y_r(T_f)$. The parameter $T_f$ is the total duration of the reference trajectory $y_r(t)$, and the time constant $\tau$ is commonly assumed to be unitary.

DMP  represent a simple yet effective encoding technique, featuring useful properties such as stability, time scaling, spatial modulation, and compatibility with additional coupling terms~\cite{4,5,10}. Thanks to their compact and versatile formulation, DMP were successfully applied to various tasks, such as tennis swing~\cite{8}, bi-manual manipulation~\cite{7}, peg-in-hole~\cite{zoe1}, among others~\cite{6,28,29,zoe3}.
Despite the several interesting researches on DMP, we identified some research branches which, to the best of our knowledge, have not been fully investigated yet,
as discussed in Sec.~\ref{methodol_and_contrib_sect}.


The remainder of this manuscript is structured as follows. An overview of the state of the art and of the issues when dealing with DMP is given in Sec.~\ref{sec2}, whereas the contributions of this work are discussed in Sec.~\ref{methodol_and_contrib_sect}. The adopted DMP formulation is discussed in Sec.~\ref{sec3} {\black together with the introduction of the proposed Spatial Sampling algorithm, representing} the core of the proposed Geometric DMP (GDMP) formulation.
%
%
Different applications of the GDMP are then proposed in the following sections. Specifically, {\black the phase optimization problem} is discussed in Sec.~\ref{Phase_Opt_DMP_section} {\black together with the comparison with other DMP approaches}, while the human-in-the-loop application is discussed in Sec.~\ref{human_in_the_loop_section} 
together with the passivity analysis. {\black The experimental stability evaluation in co-manipulation tasks is performed in Sec.~\ref{Experimental_Tests_section}, together with the experimental comparison of GDMP with other DMP approaches.}
%
%
The conclusions are given in Sec.~\ref{Conclusions_sect}.

\section{Related Works}
\label{sec2}

The state trajectories of a dynamic physical system~\cite{tebaldi2024} naturally evolve with time. However, if the considered dynamic model does not represent a physical system, the system trajectories may be desired to evolve according to a different timing law. For instance, scaling the original time by a suitable factor can enhance flexibility, offering advantageous properties like feedback linearizability~\cite{timescaling,sampei1986time}.

In the specific case of DMP, the timing law is typically governed by the Canonical System (CS) in \eqref{cs}, where the variation of the phase variable $s(t)$ can be modulated by properly tuning the parameter $\tau$, thus giving a time-scalable Transformation System (TS) in \eqref{ts1}-\eqref{ts2} and maintaining the topological equivalence of the output trajectories~\cite{5,zoe1,zoe3}. Choosing the CS in \eqref{cs} to exhibit an exponential decay ensures that the contribute of $f(s(t))$ in \eqref{ts1}
fades out when reaching the goal point $g$~\cite{4}. However, this solution inevitably leads to very high values of the weights in \eqref{lwr} when close to the final part of the reference trajectory~\cite{9}, which can be undesirable when combining multiple DMP together~\cite{6,10,29}. To cope with this issue, other types of CS were adopted, such as piece-wise linear~\cite{9} or sigmoidal-like phase profiles~\cite{kulvicius2011modified}.

The key aspect of having a timing law governed by the CS in \eqref{cs} is that the evolution of the trajectory $y(t)$ in the TS \eqref{ts1}-\eqref{ts2} can be effectively slowed down or or sped up by tuning the parameter $\tau$, without introducing any geometric variation in the resulting trajectory $y(t)$ of the DMP~\cite{5,73}. In this case, an issue arises when the trajectory is required to stop during its reproduction, as this would imply $\tau \rightarrow \infty$. This problem was noted in~\cite{8,27}, then further analyzed in~\cite{28}. In the latter reference, the scaling factor $\tau$ relies on the phase $s(t)$ and is expressed as a combination of basis functions, aiming at acquiring an optimized velocity profile for a bimanual kitchen task.

Many different formulations of the CS in \eqref{cs} can be found in the literature, in order to adopt proper timing laws for the considered applications. 
In~\cite{kramberger2018passivity}, a coupling term is used to modulate the task velocity based on the measured contact force with the environment. The task velocity can also be changed as in~\cite{27}, where non-uniform velocity changes are allowed to move along the trajectory according to the forces and torques applied to the robot. In~\cite{nemec2016speed}, Iterative Learning Control (ILC) is employed to learn an additional and optimal phase-dependent temporal scaling factor into the DMP equations, whereas an online optimization of the time duration of the task is performed in~\cite{zoe3} in order to vary the task velocity, and thus the robot velocity, as a function of the distance from the target. Other works where the timing law is modified according to external perturbation on the planned trajectory can be found in~\cite{8,zoe1,robotics12040118,papageorgiou2020control,braglia2024minj}.

{\black A key application in trajectory planning is minimizing task duration. Shaping the timing law $s(t)$ along a predefined path for tracking is commonly known as time-optimal path parameterization (TOPP). In TOPP problems, the objective is to minimize the execution time $T$ along a constrained curve, subject to first and second order constraints~\cite{lipp2014minimum}. 
TOPP relies on the path-speed separation principle, where the constraint curve serves as a geometric constraint to which any speed profile can be applied. Robotic applications often impose constraints on the minimum allowable time duration $T$ to respect the robot's kinematic limitations. In~\cite{verscheure2009time}, this problem is transformed  into a convex optimal control problem with a single variable. Alternatively, \cite{chen2024generalizing} solves the TOPP using factor graph variable elimination, achieving the global optimum for minimum time problems with quadratic objectives. }

{\black In the DMP literature, the minimum time problem has been indirectly addressed due to temporal scaling being a well-known property of DMPs. For example, scaling $\tau$ in the canonical system of~\eqref{cs} can reduce time duration, but it does not enforce kinematic constraints, such as velocity or acceleration limits. This limitation was acknowledged in~\cite{dahlin2019adaptive,dahlin2021temporal}, where the authors proposed a novel DMP formulation that scales $\tau$ offline and online to comply with the kinematic constraints. Similarly, in~\cite{zoe2}, the problem of respecting kinematic constraints was addressed by appropriately computing the weights $\omega_i$ in~\eqref{eq8}.
While these approaches effectively scale robot velocity, they cannot be classified as TOPP methods because (i) position trajectories are used instead of geometric paths, and (ii) their objective is not to minimize time duration. To the best of our knowledge, the minimum time problem has not yet been directly addressed in the DMP literature. }

Another  important aspect when dealing with timing laws is the possibility of letting the trajectory move backward and forward between the initial and goal positions, $y_0$ and $g$ respectively, while preserving the trajectory topology, i.e. while allowing $\dot{s}(t)$ to change sign~\cite{zoe1,verscheure2009time}. This feature is called phase reversibility. Even though it offers many benefits in various scenarios~\cite{27}, the classical DMP formulation in \eqref{ts1}-\eqref{cs} does not fulfill the phase reversibility requirement~\cite{nemec2016speed}. In fact, the restriction of having a strictly monotonic $\dot{s}(t)$ implies that the new timing law neither allows the trajectory to be reproduced backward nor allows it to be stopped~\cite{sampei1986time}. 
The problem of reversibility was considered in~\cite{nemec2018efficient} and further investigated in~\cite{zoe1,14}, where it was shown that having a reversible phase can affect the DMP global stability property~\cite{27}.
In~\cite{14}, the authors formulate the design requirements for guaranteeing phase reversibility in DMP, which is a desirable property in some DMP formulation~\cite{zoe1,zoe2,zoe3,zoe4,escarabajal2023imitation}.
Firstly, a reversible DMP should maintain global stability properties as the original one~\cite{14}, meaning that
$y_0$ must be an attractor point when moving backward, whereas $g$ must be an attractor point when moving forward. Secondly, the trajectory topology must be preserved when moving forward and backward. Finally, it would be desirable to adopt the same forcing term when progressing along the intended path, rather than switching between two different functions depending on the motion direction.

 Whenever a new DMP formulation is proposed, it is also important to verify that global asymptotic stability is guaranteed.
 In~\cite{14}, a candidate for such a system is given by the Logistic Differential Equation (LDE) $\tau\dot{y} = \alpha(y-y_0)(g-y)$. However, it can be demonstrated that the region of global stability of the LDE is restricted to $y \in (y_0,g)$ with  $y_0 < g$~\cite{wiggins2003introduction}.
 Consequently, two separate forcing terms are learned to switch along the desired curve whenever the sign of $\dot{s}$ changes~\cite{zhao2023robotic}.

When dealing with phase reversibility, a considerable step forward was provided in~\cite{zoe1}. In this work, the authors present a new formulation of the original DMP satisfying all phase reversibility requirements by using a unique forcing term $f(s(t))$, based on a parameterization of the reference trajectory $y_r(t)$ and of its time-derivatives ~\cite{vpsto,verscheure2009time}.
In this case, there is no need to limit the forcing term $f(s(t))$
by multiplying it by $s(t)$, as the terms $y_r(t),\dot{y}_r(t),\ddot{y}_r(t)$ are bounded by construction and the derivatives tend to zero when the system approaches the initial/goal positions~\cite{zoe2}. Additionally, backward reproduction is achieved without impacting the DMP stability by simply reversing the sign of the phase velocity $\dot{s}(t)$~\cite{zoe3}.

A potential application for reversible DMP is its use as a virtual fixture~\cite{raiola2015co,restrepo2017iterative,amor2014interaction}. 
The latter represents a predefined path constraining the robot movement without a specified timing law. The phase variable $s(t)$ can be adjusted arbitrarily, allowing the system to evolve along the predefined paths. However, the DMP architecture imposes some limitations as the recorded reference trajectories are affected by the time dependence observed during the task demonstrations~\cite{33}.
The same issue emerges when multiple demonstrations of the same task are recorded with different timing laws. In this case, techniques such as Dynamic Time Warping (DWT)~\cite{berndt1994using} allow to align the different demonstrations and compute a unique reference motion, such that the related skill is extracted~\cite{hovland1996skill,calinon2007learning}.

In this paper, we address the issues detailed in this section by proposing the concept of Geometric DMP (GDMP). This approach is based on the introduction of a Spatial Sampling algorithm, which enables a geometric interpretation of the phase variable $s(t)$. Unlike the standard case, this new phase variable is no longer constrained by the timing law of the demonstration and can be appropriately chosen based on the specific application. This characteristic justifies the alternative name we have given to this method: Phase-Independent DMP.

\section{Methodology and Contributions}\label{methodol_and_contrib_sect}

Despite the extensive literature on DMP, to the best of our knowledge, there is still no unified framework that allows for the arbitrary selection of the Canonical System in \eqref{cs}.
In this work, we aim at achieving this objective starting from the DMP formulation originally proposed in \cite{zoe1}. The new contributions of our work are:
1) The proposal of a new Spatial Sampling algorithm. This algorithm allows the conversion of the demonstrated trajectory, which is constantly time-sampled, into a new one which is constantly spatially-sampled. 
The demonstrated trajectory can {\black can also include parts with zero speed}.
The spatially-sampled trajectory
no longer exhibits the time dependence of the original one, and is expressed as a function of its arc-length parameter. 
{\black For this reason, it is a \textit{regular curve}, i.e., a curve whose first derivative is always nonzero.}
2) {\black The use of the spatial sampling algorithm allows to decouple}
the phase variable from the timing law during the demonstration phase. {\black Therefore, the term ``phase-independent'' DMP is introduced, referring to the fact that the} phase variable can now be freely chosen depending on the application and is no longer constrained by the Canonical System \eqref{cs}. {\black Time scaling and reversibility are direct consequences of this feature.}
Since the forcing term $f(s(t))$ is computed solely based on the geometrical information contained in the demonstrated trajectory, the proposed methods is also called Geometric DMP (GDMP).\\
 {\black Two significant applications of the GDMP are presented: 3) the computation of the phase variable that optimizes a given cost function, such as total execution time, subject to velocity and acceleration constraints; 4) a human-robot co-manipulation task, where the robot constrains the motion along the path encoded by the GDMP, while the human determines how the motion unfolds. As a feedback system with the human in the loop, a passivity analysis and an experimental evaluation of practical stability are provided. Both applications are compared with several approaches from the DMP literature, demonstrating the enhanced performance of GDMP.
}
\begin{figure}[t]
\centering \small \setlength{\unitlength}{2.6mm}
\psset{unit=\unitlength} \SpecialCoor
\begin{pspicture}(1,-6)(30,15.5)
\rput(0.5,3){$s(t)$}
\psline{->}(1.5,3)(2.5,3)(2.5,7)(3.5,7)
\psline{->}(2.5,3)(3.5,3)
\psline{->}(2.5,3)(2.5,-1)(3.5,-1)
\rput(-0.5,0){
\psline[linearc=.2,linecolor=black](4.5,-2.25)(7.5,-2.25)(7.5,0.25)(4,0.25)(4,-2.25)(4.5,-2.25)
\psline[linearc=.2,linecolor=black]{->}(7.5,-1)(8.5,-1)
\pscircle[](9,-1){.5} \psline[linecolor=black,linewidth=0.4pt](8.6,-1)(8.9,-1)
\psline[linecolor=black,linewidth=0.4pt](8.75,-1.15)(8.75,-0.85)
\psline[linecolor=black,linewidth=0.4pt](8.85,-1.25)(9.15,-1.25)
\psline[linearc=.2,linecolor=black]{->}(9,-3)(9,-1.5)

\psline(8.25,-3)(9.75,-3)(9.75,-4.5)(8.25,-4.5)(8.25,-3)
\rput(9,-3.75){$g_r$}
\psline[linearc=.2,linecolor=black]{->}(9.5,-1)(11.25,-1)
\psline[linecolor=black](11.25,-1)(11.25,-2)(13.25,-1)(11.25,0)(11.25,-1)
\rput(12,-1.05){\tiny $\alpha \beta$}
\psline[linecolor=black]{->}(13.25,-1)(15,-1)(15,1)
\rput(0,4){
\psline[linearc=.2,linecolor=black](4.5,-2.25)(7.5,-2.25)(7.5,0.25)(4,0.25)(4,-2.25)(4.5,-2.25)
\rput(5.75,-1){\scriptsize $y^{\star\prime}(\!s(t)\!)$}}
\psline[linecolor=black](7.5,3)(9,3)(9,4.5)(9.75,4.5)
\psline[linearc=.2,linecolor=black,dash=.8pt,linestyle=dashed](9.75,4.5)(10.75,4.5)
\psline[linecolor=black]{->}(10.75,4.5)(11.25,4.5)
\pscircle[](11.75,4.5){.5} \rput(11.75,4.5){$\times$}
\psline[linecolor=black]{->}(9,3)(9,1.5)(9.75,1.5)
\pscircle[](10.25,1.5){.5} \rput(10.25,1.5){$\times$}
\psline[linecolor=black]{->}(10.75,1.5)(11.75,1.5)
\psline[linecolor=black](11.75,1.5)(11.75,0.5)(13.75,1.5)(11.75,2.5)(11.75,1.5)
\rput(12.5,1.5){\tiny $\alpha$}
\psline[linecolor=black]{->}(12.25,4.5)(13,4.5)(13,6.5)
\psline[linecolor=black]{->}(13.75,1.5)(14.5,1.5)
\rput(6,2.5){
\pscircle[](9,-1){.5} \psline[linecolor=black,linewidth=0.4pt](8.6,-1)(8.9,-1)
\psline[linecolor=black,linewidth=0.4pt](8.75,-1.15)(8.75,-0.85)
\psline[linecolor=black,linewidth=0.4pt](8.85,-1.25)(9.15,-1.25)
\psline[linecolor=black,linewidth=0.4pt](9,-1.4)(9.,-1.1)%
}
\psline[linecolor=black]{->}(15,2)(15,6.5)

\rput(0,8){
\psline[linearc=.2,linecolor=black](4.5,-2.25)(7.5,-2.25)(7.5,0.25)(4,0.25)(4,-2.25)(4.5,-2.25)
\rput(5.75,-1){\scriptsize $y^{\star\prime\prime}\!(\!s(t)\!)$}}
\rput(5.75,-1){\scriptsize $y^\star\!(\!s(t)\!)$}
\psline[linearc=.2,linecolor=black]{->}(7.5,7)(8.5,7)
\pscircle[](9,7){.5} \rput(9,7){$\times$}
\psline[linearc=.2,linecolor=black](9.5,7)(9.75,7)
\psline[linearc=.2,linecolor=black,dash=.8pt,linestyle=dashed](9.75,7)(10.75,7)
\psline[linearc=.2,linecolor=black](10.75,7)(11,7)
\rput(1.5,0){
\psline[linearc=.2,linecolor=black](9.5,7)(9.75,7)
\psline[linearc=.2,linecolor=black,dash=.8pt,linestyle=dashed](9.75,7)(10.75,7)
\psline[linearc=.2,linecolor=black](10.75,7)(11,7)}
\rput(4,8){
\pscircle[](9,-1){.5} \psline[linecolor=black,linewidth=0.4pt](8.6,-1)(8.9,-1)
\psline[linecolor=black,linewidth=0.4pt](8.75,-1.15)(8.75,-0.85)
\psline[linecolor=black,linewidth=0.4pt](8.6,-1)(8.9,-1)
\psline[linecolor=black,linewidth=0.4pt](8.85,-1.25)(9.15,-1.25)
\psline[linecolor=black,linewidth=0.4pt](9,-1.4)(9.,-1.1)}
\psline[linearc=.2,linecolor=black](13.5,7)(14.5,7)
\rput(6,8){
\pscircle[](9,-1){.5} \psline[linecolor=black,linewidth=0.4pt](8.6,-1)(8.9,-1)
\psline[linecolor=black,linewidth=0.4pt](8.75,-1.15)(8.75,-0.85)
\psline[linecolor=black,linewidth=0.4pt](8.6,-1)(8.9,-1)
\psline[linecolor=black,linewidth=0.4pt](8.85,-1.25)(9.15,-1.25)
\psline[linecolor=black,linewidth=0.4pt](9,-1.4)(9.,-1.1)}

\psline{->}(9,8.25)(9,7.5)
\rput(-4,0){\psline[linearc=.2](13,8.25)(14,8.25)(14,10.25)(12,10.25)(12,8.25)(13,8.25)
\rput(13,9.25){\scriptsize ${(\cdot)}^2$}}
\psline{->}(2,11)(9,11)(9,10.25)
\rput(1,11){$\dot{s}(t)$}
\psline{->}(9,11)(10.25,11)(10.25,2)
\psline{->}(2,14)(11.75,14)(11.75,5)
\rput(1,14){$\ddot{s}(t)$}

}
\rput(15.85,8.5){\scriptsize $f^{\!\star}(\!s(t)\!)$}
\psline{->}(15,7)(16,7)
\psline(16,7)(16,8)(18,7)(16,6)(16,7)
\rput(16.75,7){\tiny $\eta$}
\psline{->}(18,7)(19,7)

\rput(10.5,8){
\pscircle[](9,-1){.5} \psline[linecolor=black,linewidth=0.4pt](8.6,-1)(8.9,-1)
\psline[linecolor=black,linewidth=0.4pt](8.75,-1.15)(8.75,-0.85)
\psline[linecolor=black,linewidth=0.4pt](8.85,-1.25)(9.15,-1.25)
\psline[linecolor=black,linewidth=0.4pt](9,-1.4)(9.,-1.1)%
}
\rput(1,0){
\psline[linecolor=black]{->}(19,7)(20.135,7)
\rput(7.9,-2.25){\psline[linearc=.2,linecolor=black](13,8.25)(14.12,8.25)(14.12,10.25)(12.12,10.25)(12.12,8.25)(13,8.25)}
}
\rput(20,8){\scriptsize $\ddot{y}(t)$}
\rput(22.04,7){\scriptsize $\int $}
\psline[linecolor=black]{->}(23.06,7)(25.25,7)
\rput(5.5,0){
\rput(7.25,-2.25){\psline[linearc=.2,linecolor=black](13,8.25)(14.44,8.25)(14.44,10.25)(12.44,10.25)(12.44,8.25)(13,8.25)}
\rput(18.8,8){\scriptsize $\dot{y}(t)$}
\rput(20.69,7){\scriptsize $\int $}}
\psline[linecolor=black]{->}(27.24,7)(30,7)
\rput(1.5,0){
\psline{->}(22.5,7)(22.5,4)(21.48,4)
\psline(21.48,4)(21.48,5)(19.48,4)(21.48,3)(21.48,4)
\rput(20.68,4){\tiny $\alpha$}
}
\psline{->}(20.98,4)(20,4)
\rput(10.5,5){
\pscircle[](9,-1){.5}
\rput(0.25,-0.25){
\psline[linecolor=black,linewidth=0.4pt](8.6,-1)(8.9,-1)
}
\rput(0.25,0.25){
\psline[linecolor=black,linewidth=0.4pt](8.85,-1.25)(9.15,-1.25)
}
}
\psline[linecolor=black]{->}(19.5,4.5)(19.5,6.5)
%
\psline[linecolor=black]{->}(27.5,1)(24.68,1)
\rput(2.5,0){
\rput(1.2,-3){\psline(21,4)(21,5)(19,4)(21,3)(21,4)
\rput(20.168,4){\tiny $\alpha\beta$}}
}
\psline[linecolor=black]{->}(22.8,1)(19.5,1)(19.5,3.5)

\rput(19,2){
\pscircle[](9,-1){.5}
\rput(0.25,0.25){
\psline[linecolor=black,linewidth=0.4pt](8.6,-1)(8.9,-1)
\psline[linecolor=black,linewidth=0.4pt](8.75,-1.15)(8.75,-0.85)}
\rput(0.25,0.25){
\psline[linecolor=black,linewidth=0.4pt](8.85,-1.25)(9.15,-1.25)}}
\psline{->}(28,7)(28,1.5)
\psline{->}(30,1)(28.5,1)
\rput(31,7){$y(t)$}
\rput(30.8,1){$g$}

\rput(1.2,0){
\psline[linecolor=myorangenew,linestyle=dashed](0.8,-2.5)(14.1,-2.5)(14.1,14.5)(0.8,14.5)(0.8,-2.5)}
\rput(6.4,12.5){\textcolor{myorangenew}{Eq.~\eqref{eq13}}}
\rput(-0.5,0){
\psline[linecolor=mygreennew,linestyle=dashed](18.86,-2.5)(29.75,-2.5)(29.75,9)(18.86,9)(18.86,-2.5)}
\rput(24,-1){\textcolor{mygreennew}{Computation of $y(t)$}
}
%
\end{pspicture}
\vspace{-4.6mm}
\caption{Block-scheme representation of the TS \eqref{eq:SecondOrderSystem} by incorporating the forcing term \eqref{eq13}, which ensures the independence from the phase variable $s(t)$.}\label{fig:DMPnewScheme}
\end{figure}
\vspace{-3mm}

\begin{figure*}
\psfrag{x}[c][t][0.9]{$x$ [m] }
\psfrag{y}[c][t][0.9]{$y$ [m] }
\psfrag{z}[c][t][0.9]{$z$ [m] }
\psfrag{data111}[][][0.68]{$y(t)$}
\psfrag{data222}[][][0.68]{$y_1(t)$}
\psfrag{data333}[][][0.68]{$y_2(t)$}
\psfrag{dat4}[][][0.68]{$y_r(t)$}
\psfrag{g1}[c][t][0.6]{$\g_r$}
\psfrag{g2}[c][t][0.6]{$\g_1$}
\psfrag{g3}[c][t][0.6]{$\g_2$}
\psfrag{y0}[c][t][0.6]{$\y_r(0)$}
\psfrag{data999}[][][0.68]{$y(t)$}
\psfrag{data110}[][][0.68]{$y_1(t)$}
\psfrag{data133}[][][0.68]{$y_2(t)$}
\psfrag{dat5}[][][0.68]{$y_r(t)$}
\psfrag{y1}[c][t][0.6]{$\y_r(0)$}
\psfrag{y2}[c][t][0.6]{$\!\!\!\!\!\!\!\!\!\!\!\!\y_1(s(0))$}
\psfrag{y3}[c][t][0.6]{$\!\!\!\!\!\!\!\!\y_2(s(0))$}
\psfrag{g0}[c][t][0.6]{$\g_r$}

\psfrag{obs1}[][][0.6]{\small $\;$Obj1}
\psfrag{obs2}[][][0.6]{\small $\;$Obj2}
\psfrag{y00}[c][t][0.7]{\hspace{-12.5mm}$\y_r(0)$ }
\psfrag{g00}[c][t][0.7]{ \hspace{3mm}$\g_r$ }
\psfrag{ref}[][][0.68]{$\!\!\!y_r(t)$}
\psfrag{gdmp}[b][b][0.68]{$y(t)$}

\centering
\begin{subfigure}{0.3\linewidth}
\centering
    \includegraphics[width=\columnwidth]{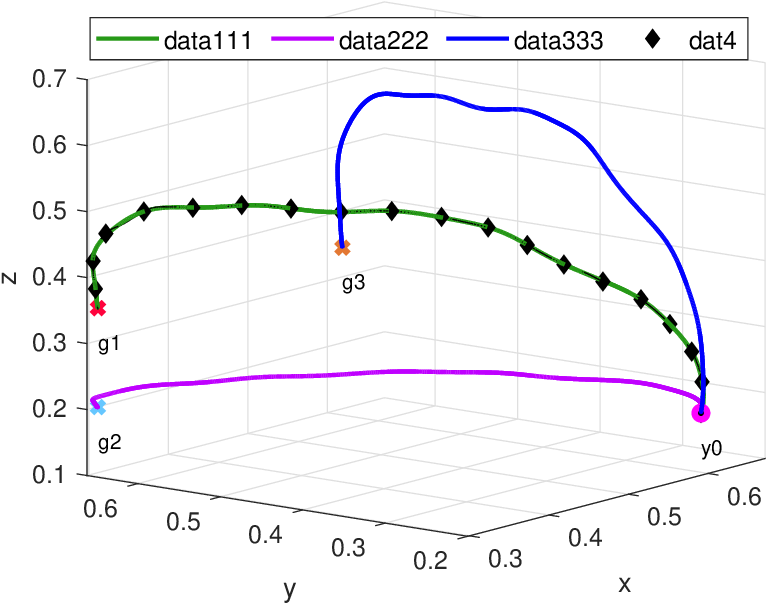} 
    \caption{}
    \label{subfig:g}
\end{subfigure}\hspace{0.3cm}
\begin{subfigure}{0.3\linewidth}
\centering
    \includegraphics[width=\columnwidth]{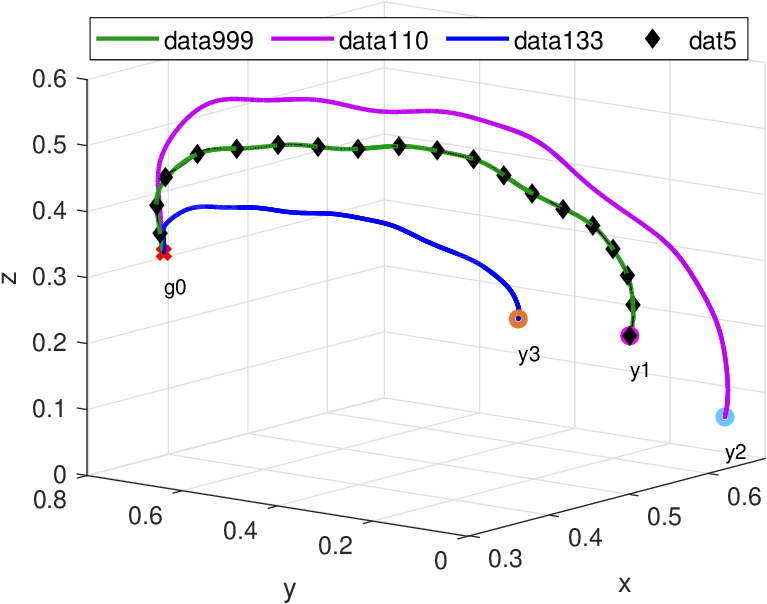} 
    \caption{}
    \label{subfig:y0}
\end{subfigure}\hspace{0.3cm}
\begin{subfigure}{0.3\linewidth}
\centering
    \includegraphics[width=\columnwidth]{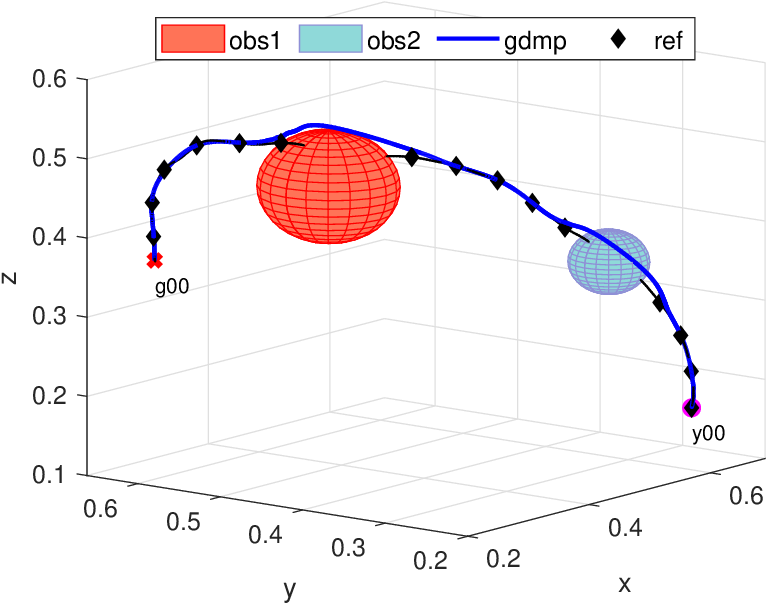} 
    \caption{}
    \label{subfig:coupling}
\end{subfigure}
    \caption{Verification that the DMP formulation in Fig.~\ref{fig:DMPnewScheme} maintains the characteristic properties of the original one: spatial adaptation with respect to (a) different target goals, (b) different initial positions and (c) presence of obstacles.}
\label{fig:gdmp_properties}
\end{figure*}

\section{Geometric DMP Formulation}
\label{sec3}
{\black This section deals with the description of the proposed Geometric DMP, which employs the formulation in \cite{zoe1} as a starting point.}
The forcing term $f(s(t))$ in \eqref{eq:SecondOrderSystem} is computed as follows:
\begin{equation}
f(s(t)) = \eta f^{\star}(s(t)), \;\;\; \mbox{where}
\;\;\;     \eta = \frac{ g-y(s(0)) }{ g_r-y_r(0) }
\label{eq:ForcingTerm}
\end{equation}
is a scaling factor depending on the initial and goal positions $y_r(0)$ and $g_r = y_r(s_{f})$ of the reference trajectory, as well as on the initial and desired goal positions $y(s(0))$ and $g$.
The objective of the scaling factor $\eta$ is to preserve the topology of the curve~\cite{73,zoe3,zoe4}.

In the classical DMP, the forcing term $f^{\star}(s)$ is obtained as in \eqref{fs}, \eqref{rbf} and \eqref{lwr}, where the nominal forcing term $f_r(t)$ is computed as a linear combination of the demonstrated position, velocity, and acceleration profiles as in \eqref{fd}. In our case, 
the forcing term $f^{\star}(s)$ in \eqref{eq:ForcingTerm} is obtained using the linear combination \eqref{fd} by replacing the demonstrated position, velocity, and acceleration profiles with the 
corresponding parameterizations 
$y^\star(s)$, $\dot y^\star(s)$ and $\ddot y^\star(s)$. Specifically, given the sequence of demonstrated positions   $y_r(s)$, $s=0,\ldots,s_{f}$, the coefficients $\omega_i$ of the parametric function
\begin{equation}
\label{eq8}
   y^{\star}(s) =
  \sum_{i=1}^{N} \omega_i \phi_i(s)
\end{equation}
are determined by minimizing the cost function
\begin{equation}
\label{eq9}
    J = \sum_{t=0}^{T_f} \left\|  y_r(s) - \omega_i\phi_i(s) \right\|^2.
\end{equation}
The radial basis functions $\phi_i(\cdot)$ are defined in \eqref{rbf}. In \cite{zoe1}, the independent variable is the time of the demonstrated trajectory, i.e. $s\equiv t$.  The function $y^{\star}(s)$ is then reparameterized by composing it with the desired timing law $s = s(t): [0, T_{f}]\rightarrow [0, s_{f}]$. Note that $s(t)$ must be a monotonic  $C^{1}$-class function such that $\dot s(0) = \dot s(T_{f}) =0$ and $\ddot s(0) = \ddot s(T_{f}) =0$.
The derivatives $\dot{y}^{\star}(s(t))$ and $\ddot{y}^{\star}(s(t))$ of the parameterized trajectory $y^{\star}(s(t))$ can be deduced as follows:
\begin{equation}
\label{eq16}
\begin{array}{@{\!}r@{\,}c@{\,}l}
    \dot{y}^{\star}(s(t)) &=& \dfrac{\partial y^{\star}(s(t))}{\partial s(t)} \dfrac{ds(t)}{ dt} = y^{\star\prime}(s(t))\dot{s}(t),
     \\[4mm]
         \ddot{y}^{\star}(s(t)) & = &\dfrac{d \dot{y}^{\star}(s(t))}{dt}=
         y^{\star\prime\prime}(s(t))\dot{s}^2(t) \!\!+\!y^{\star\prime}(s(t))\ddot{s}(t),
\end{array}
\end{equation}
where the prime and dot notations denote the differentiation with respect to the phase variable $s(t)$ and with respect to time, respectively. Therefore, with the proposed formulation, it is not necessary to know the derivatives of the reference trajectory, but only the position profile $y_r(t)$ is sufficient.
Using  \eqref{eq:ForcingTerm}, \eqref{eq8} and \eqref{eq16}, the forcing term $f(s(t))$ in 
\eqref{eq:SecondOrderSystem} can be written as:
\begin{eqnarray}
f(s(t))&\!\!\!\!\!=\!\!\!\!\!& \eta \Big[y^{\star\prime\prime}(s(t))\dot{s}^2(t) \!+y^{\star\prime}(s(t))\ddot{s}(t)   \nonumber \\ && +\alpha y^{\star\prime}(s(t))\dot{s}(t) +  \alpha\beta \big( y^{\star}(s(t)) -g_r\big)\Big]. \label{eq13}
\end{eqnarray}
By distributing the scaling factor $\eta$ across all the terms on the right side of \eqref{eq13}, $f(s(t))$ can be expressed as a linear combination of
\begin{equation}
y^\star_\eta(s(t)) = \eta\, y^{\star}(s(t))
\label{eq:y_eta}
\end{equation}
and its derivatives.
The complete schematic view of the TS \eqref{ts1}-\eqref{ts2} with the forcing term \eqref{eq13} is reported in Fig.~\ref{fig:DMPnewScheme}, and enables the following features~\cite{zoe1,zoe2}: time modulation by varying the evolution of the phase variable $s(t)$; phase reversibility allowing forward and backward reproduction of the trajectory, when $\dot s (t)>0$ and $\dot s (t)<0$, respectively; spatial scaling of the curve varying the initial position $y(s(0))$  and the goal $g$ of the task.

The TS can then be generalized to the multidimensional case. If the considered application requires the robot position only to be considered, as in Sec.~\ref{Phase_Opt_DMP_section}, Sec.~\ref{human_in_the_loop_section} and Sec.~\ref{Experimental_Tests_section}, a Cartesian representation of the TS can be employed:
\begin{equation}\label{DMP_m_star}
\begin{array}{r@{\,}c@{\,}l}
\ddyorig + \A \dyorig +
\A\B\big(\yorig-\g
\big)
& =&\E\f^{\star}(s(t)),
\end{array}
\end{equation}
\noindent where $\A\!=\!\mbox{diag} \left(\alpha_x,\alpha_y,\alpha_z\right)$, $\B\!=\!\mbox{diag} \left(\beta_x,\beta_y,\beta_z\right)$, $\E\!=\!\mbox{diag} \left(\eta_x,\eta_y,\eta_z\right)$, and
\begin{equation}
\label{DMP_m1_star}
  \yorig\!=\!\mat{@{}c@{}}{y_{x}(t) \\[1mm]
 y_{y}(t) \\[0.2mm]
 y_{z}(t) },\,\,\g\!=\!\mat{@{}c@{}}{g_{x} \\[1mm]
 g_{y} \\[0.2mm]
 g_{z} },\,\, \f^{\star}(s)\!=\!\mat{@{}c@{}}{f^{\star}_x(s(t)) \\[1mm]
 f^{\star}_y(s(t)) \\[1mm]
 f^{\star}_z(s(t)) }.
\end{equation}
Note that Eqs.~\eqref{DMP_m_star}-\eqref{DMP_m1_star}can also be generalized to account for the end-effector orientation by adopting a minimal representation, such as Euler angles or unit quaternions \cite{5,zoe4,koutras2020correct}.
The resulting TS in Fig.~\ref{fig:DMPnewScheme} still maintains the characteristic properties of the original formulation, such as spatial adaptation with respect to different goal/initial position and with respect to the presence of obstacles by adding proper coupling terms~\cite{5}. Figure~\ref{fig:gdmp_properties} shows an example of application of such properties: Fig.~\ref{fig:gdmp_properties}a and Fig.~\ref{fig:gdmp_properties}b show the possibility of changing
the goals and initial positions $g$ and $y(s(0))$ in \eqref{eq:ForcingTerm} with respect to the reference ones $g$ and $y_r(0)$, while Fig.~\ref{fig:gdmp_properties}c shows how the introduction of coupling terms, in this case repulsive potential fields~\cite{gams2016adaptation,zoe2}, allows to avoid obstacles along the path.

{\black 
\subsection{Regularity of the Curve}\label{regularity_section}
}

The Canonical System (CS) in~\eqref{cs} can be set in order to ensure that the output of the DMP accurately reproduces the reference trajectory, including its time dependency~\cite{8}.
This means that, if the user stops while recording the reference trajectory, i.e. $\frac{d{\y}_r{(s)}}{ds}=0$, the corresponding parameterized trajectory will result in $\y^{\star\prime}(s)=0$.
In all the formulation of DMP proposed in the literature, the duration of this pausing phase can be shortened or extended by adjusting the parameter $\tau$ in~\eqref{ts1}-\eqref{cs}, but cannot be removed~\cite{restrepo2017iterative}. 
{\black This side effect persists also in the formulation of Fig.~\ref{fig:DMPnewScheme} 
proposed in~\cite{zoe1}, since $\dot{\y}_r{(t)}=0$ implies $\dot \y^\star(s(t)) = \ddot \y^\star(s(t)) =0 $. 
This issue becomes particularly critical in applications where only the geometric aspects of the demonstrated path are required. For instance, in~\cite{zoe1}, the reference trajectory is captured in two distinct steps: one for the path and another for the velocity profiles. Similarly, in~\cite{27}, a dynamic scaling factor is employed, modulated by the tangential component of an external force. }

{\black The aforementioned problem can be formally identified as the lack of regularity in the curve. By definition~\cite{toponogov2006differential}, a curve $\hat \y_r(s): \mathds{R} \rightarrow \mathds{R}^3$ is regular iff:
\begin{equation}\label{regul_eq}
\dfrac{d \hat \y_r(s)}{d s} \neq 0 \hspace{8mm} \forall s \in \mathds{R}.
\end{equation}
From \eqref{regul_eq}, it follows that a regular curve $\hat \y_r(s)$ and its tangential direction$\frac{d \hat \y_r(s)}{d s}$ are always well-defined at each point of the space. 
}

\vspace{.5cm}
\subsection{Spatial Sampling Algorithm}

To address the aforementioned issues, we propose re-sampling the recorded trajectory  $\y_r(t)$ using the proposed ``Spatial Sampling'' algorithm as a preliminary step to the parameterization \eqref{eq8}. The proposed algorithm generates a filtered trajectory $\yf(\!\sfk\!)$ which is a function of the arc-length parameter 
$\sfk$. Other approaches have been proposed in the literature to address the separation of the spatial and temporal components in the demonstrated trajectory, such as the arc-length (AL) DMP proposed in~\cite{GASPAR2018225}. However, the ALDMP proposed in~\cite{GASPAR2018225} cannot handle situations where the user stops while recording the demonstrated trajectory, namely it cannot handle situations where $\dot{\y}_r(t)=0$ without introducing a trajectory segmentation.
\begin{algorithm}[t] 
{\black
 \caption{{\black
Spatial Sampling}}
 \begin{algorithmic}[1]
 \renewcommand{\algorithmicrequire}{\textbf{Input: $\Y_T \in \mathds{R}^{(N\times d)}, 
 \; \t_T \in \mathds{R}^{(N\times 1)}, 
 \; \Delta$}} 
 \renewcommand{\algorithmicensure}{\textbf{Output: $   \Y_{\Delta} \in \mathds{R}^{(M\times d)},\;
 \s_{\Delta} \in \mathds{R}^{(M\times 1)}, \; \t_{\Delta} \in \mathds{R}^{(M\times 1)}$} $\hspace{2cm}$}
 \REQUIRE 
 \ENSURE  
 \textit{Initialization}: \\ 
 \STATE Initialize arrays: $\!\Y_{\Delta,0} \!= \!\Y_{T,0}$, $\t_{\Delta,0} \!= \!\boldsymbol{0}$,  $\s_{\Delta,0} \!= \!0$
 \STATE Initialize current state: $\y_c \!= \! \Y_{T,0}$, $i=0$
 \\ \textit{Compute $ \Y_{\Delta},\;
 \t_{\Delta},
 \; \s_{\Delta}$} : \\
 \WHILE{$i<N$}
 %
 %
  \IF[]{ $|| \Y_{T,i+1} - \y_c|| > \Delta$}
  \STATE Compute $\Y_{\Delta,k}$ belonging to the segment $ \overline{\Y_{T}(i)\Y_{T}(i+1)} $ and such that $|| \Y_{\Delta,k} - \y_c|| = \Delta $\\
  \STATE Compute corresponding $\t_{\Delta,k}$ as in~\eqref{Spatial_sequence}
  \STATE Update arrays: $\Y_{\Delta} \!=\! [\Y_{\Delta}, \Y_{\Delta,k}]$, $\s_{\Delta} \!=\! [\s_{\Delta}, \s_{\Delta,end}+\Delta]$, $\t_{\Delta} \! = \! [\t_{\Delta}, \t_{\Delta,k}$]
  \STATE Update current state: $\y_c \!=\! \Y_{\Delta,end}$
  \ELSE 
  \STATE $i=i+1$
  \ENDIF
 \ENDWHILE
 \RETURN $   \Y_{\Delta},\;\s_{\Delta},\;
 \t_{\Delta}$ \end{algorithmic} 
\label{algorithm:SS}
}
\end{algorithm}
\begin{figure}[t!]
  \centering
  \psfrag{t}[t][t][1]{$t$}
  \psfrag{tr1}[t][t][0.68]{$\!\! \trscaluno$}
  \psfrag{tr2}[t][t][0.68]{$\!\!\!\! \trscaldue$}
  \psfrag{tr3}[t][t][0.68]{$\!\! \trscaltre$}
  \psfrag{tr4}[t][t][0.68]{$\!\! \trscalquattro$}
  \psfrag{tr5}[t][t][0.68]{$\!\! \trscalcinque$}
  \psfrag{tr6}[t][t][0.68]{$\!\! \trscalsei$}
  \psfrag{tr7}[t][t][0.68]{$\!\! \trscalsette$}
  \psfrag{tr8}[t][t][0.68]{$\!\! \trscalotto$}
  \psfrag{tr9}[t][t][0.68]{$\!\! \trscalnove$}
  \psfrag{tr10}[t][t][0.68]{$\!\! \trscaldieci$}
  \psfrag{tr11}[t][t][0.68]{$\!\! \trscalundici$}
  \psfrag{tr12}[t][t][0.68]{$\!\! \trscaldodici$}
  \psfrag{tr13}[t][t][0.68]{$\!\! \trscaltredici$}
   \psfrag{tr14}[t][t][0.68]{$\!\! \trscalquattordici$}
   \psfrag{tr15}[t][t][0.68]{$\!\! \trscalquindici$}
  \psfrag{tf1}[t][t][0.68]{$\!\! \tfscaluno$}
  \psfrag{tf2}[t][t][0.68]{$\!\! \tfscaldue$}
  \psfrag{tf3}[t][t][0.68]{$\!\! \tfscaltre$}
  \psfrag{tf4}[t][t][0.68]{$\!\! \tfscalquattro$}
  \psfrag{tf5}[t][t][0.68]{$\!\! \tfscalcinque$}
  \psfrag{tf6}[t][t][0.68]{$\!\! \tfscalsei$}
  \psfrag{tf7}[t][t][0.68]{$\!\!\!\! \tfscalsette$}
  \psfrag{tf8}[t][t][0.68]{$\!\!\! \tfscalotto$}
  \psfrag{tf9}[t][t][0.68]{$\, \tfscalnove$}
  \psfrag{yf}[b][b][1]{$\hspace{3cm}{\small \red \ylscal(t)},\,\small \yl(\tfscalk)$ } 
  \psfrag{sf}[b][b][1]{$\small \hspace{1.1cm}\sfk$}
  \psfrag{yr}[b][b][1]{$\small \hspace{2.5cm}{\red \yrscal(t)},\,\ylscal(t)$}
  \psfrag{>D}[b][b][0.68]{\textcolor{red}{$\!\!\!\!\!\!>\!\Delta$}}
  \psfrag{=D}[b][b][0.68]{\textcolor{mygreennew}{$\!\!\!\!\!\!=\!\Delta$}}
  \psfrag{<D}[b][b][0.68]{\textcolor{myorangenew}{$\;\;\;\;<\!\Delta$}}
  \psfrag{>D1}[b][b][0.68]{\textcolor{blue}{$\!\!\!\!>\!\Delta$}}
  \psfrag{D}[b][b][0.68]{\textcolor{mygreynew}{\small $\!\!\!\!\Delta$}}
  \psfrag{tt1}[t][t][0.68]{$\!\! \tfscaluno$} 
  \psfrag{tt2}[t][t][0.68]{$\!\!\!\!\!\tfscaldue$}
  \psfrag{tt3}[t][t][0.68]{$\!\!\! \tfscaltre$}
  \psfrag{tt4}[t][t][0.68]{$\!\! \tfscalquattro$}
  \psfrag{tt5}[t][t][0.68]{$\, \tfscalcinque$}
  \psfrag{tt6}[t][t][0.68]{$\!\! \tfscalsei$}
  \psfrag{tt7}[t][t][0.68]{$\, \tfscalsette$}
  \psfrag{tt8}[t][t][0.68]{$\!\! \tfscalotto$}
  \psfrag{tt9}[t][t][0.68]{$\!\! \tfscalnove$}
  \psfrag{dd}[t][t][0.68]{$\ldots$}
  \psfrag{xt1}[t][t][0.68]{$\!\!\!\!\!\!\!\!\!\! s_0$}
  \psfrag{xt2}[t][t][0.68]{$\!\!\!\!\!\!\!\!\!\! s_1$}
  \psfrag{xt3}[t][t][0.68]{$\!\!\!\!\!\!\!\!\!\! s_2$}
  \psfrag{xt4}[t][t][0.68]{$\!\!\!\!\!\!\!\!\!\! s_3$}
  \psfrag{xt5}[t][t][0.68]{$\!\!\!\!\!\!\!\!\!\! s_4$}
  \psfrag{xt6}[t][t][0.68]{$\!\!\!\!\!\!\!\!\!\! s_5$}
  \psfrag{xt7}[t][t][0.68]{$\!\!\!\!\!\!\!\!\!\! s_6$}
  \psfrag{xt8}[t][t][0.68]{$\!\!\!\!\!\!\!\!\!\! s_7$}
  \psfrag{xt9}[t][t][0.68]{$\!\!\!\!\!\!\!\!\!\! s_8$}
  \psfrag{dv}[t][t][0.68]{$\!\!\!\!\!\!\!\! \vdots$}
  %
   \psfrag{a}[b][b][0.85]{\;\;\;\;(a)}
   \psfrag{b}[b][b][0.85]{\;\;\;\;(b)}
   \psfrag{c}[b][b][0.85]{\;\;\;\;(c)}
   \psfrag{d}[b][b][0.85]{\;\;\;\;(d)}
  \includegraphics[clip,width=\columnwidth]{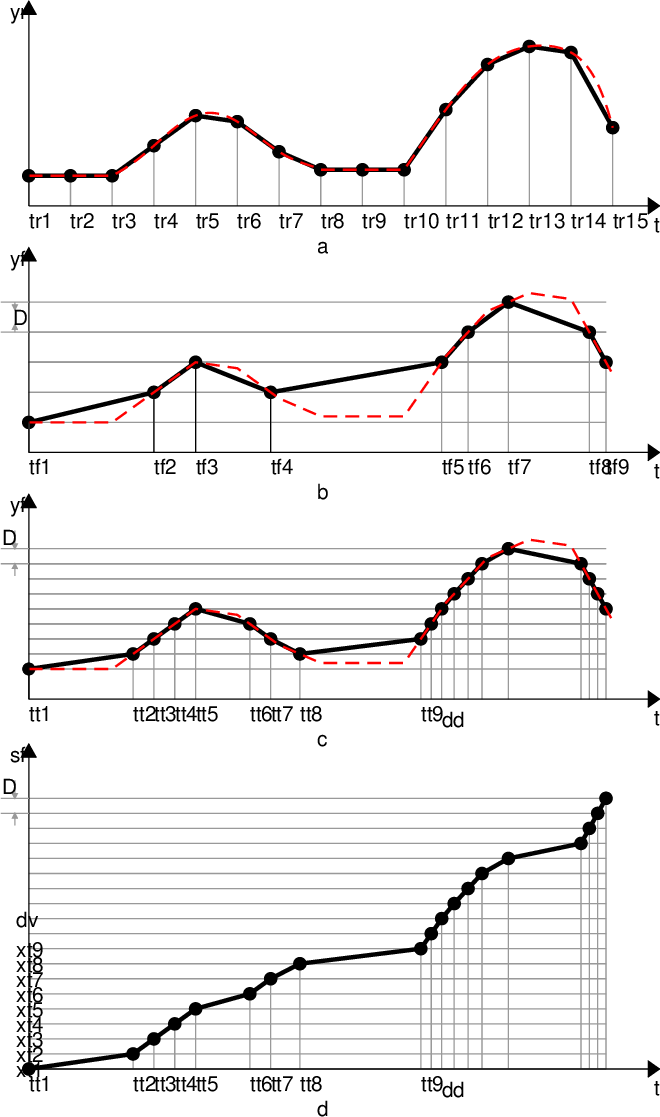}
  \psfrag{D}[b][b][0.68]{\textcolor{mygreynew}{$\!\!\!\!\Delta$}}
 \caption{Sketch displaying the operation of the Spatial Sampling algorithm in a one-dimensional scenario: reference trajectory $\ylscal(\tr)$ sampled with constant period $T$ and interpolating linear curve $\ylscal(t)$ (a), sequence of spatially sampled points   $\y_{\Delta,k}=\yl(\tfscalk)$ with $\Delta=\bar \Delta$ (b) and $\Delta=\bar \Delta/2$ (c), and samples of the function $s_k = \gamma(\tfscalk)$ for $\Delta=\bar \Delta/2$ (d).}\label{SpSa_1D_example_05_1}
\end{figure}
%

%

%
%
The proposed Spatial Sampling 
algorithm, detailed in Algorithm~\ref{algorithm:SS}, works as follows:

\noindent 1) Starting from the sequence of samples obtained by recording the demonstrated 
trajectory $\y_r(t)$ 
with constant period $T$,
 $\y_{T,k}\!=\!\yr(t_{k})\!=\!\yr(kT)  \;\; \mbox{for}\;\; k = 0, \ldots, N,$
a linearly interpolating, continuous-time function $\yl(t)$ is built by applying a First-Order Hold (FOH) on it~\cite{zeroh}. 
If the sampling period $T$ is small enough, then the following holds true:
\begin{equation}\label{new_eqq_2}
    \yl(t)\approx \yr(t).
\end{equation}

\noindent 2) A new sequence  $ \y_{\Delta,k}$ is obtained by imposing that $\y_{\Delta,0} =  \yl(0)$ and, for $k>0$, $\y_{\Delta,k} =  \yl(\tfscalk)$, where $\tfscalk$ is the time value that guarantees
\begin{equation}\label{norm_Delta1}
\| \y_{\Delta,k}-\y_{\Delta,k-1} \| =\Delta
, \;\;\; \mbox{for} \;\;\; k=1,\ldots,M.
\end{equation}
 The parameter $\Delta$ defines the geometric distance between consecutive samples and can be freely chosen.
  If the reference trajectory $\y_r(t)$ also includes the orientation components, the enforcement of condition \eqref{norm_Delta1} is restricted to the position components, while the resulting time instants $\tfscalk$ will then be used to generate the new sequence $ \y_{\Delta,k}$ including both position and orientation components.

\begin{figure*}[t]


\includegraphics[width=2\columnwidth]{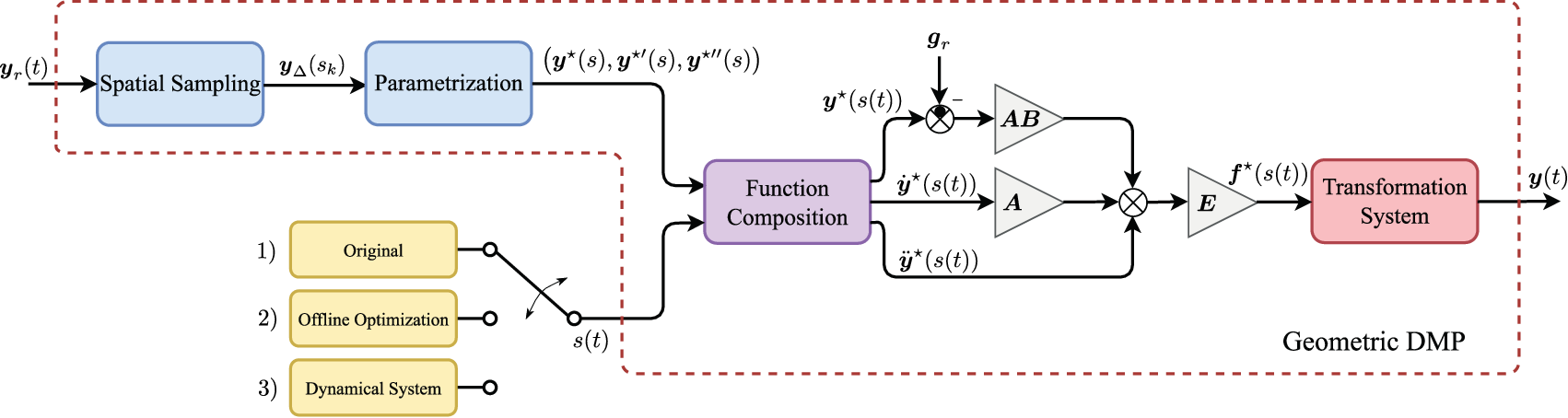}
    \caption{Schematic of the proposed framework based on Geometric DMP.}
\label{schematico}
\end{figure*}
The condition \eqref{norm_Delta1}
implies that the total distance between the first point $\y_{\Delta,0}$ and the generic $k$-th point $\y_{\Delta,k}$, $k>0$, is given by  $k\Delta$. This distance approximates the length of the curve $\yl(t)$ at the time instant  $\tfscalk$, with a precision that depends on the value of $\Delta$. Therefore, for $\Delta$ small enough, the new sampling mechanism induces a mapping between  the length $s_k = k\Delta$ and the position along the  approximating linear curve $\yl$, i.e.
\begin{equation}
\label{Spatial_sequence}
 \y_{\Delta,k}=\yl(\tfscalk), \;\;\;\mbox{ with }\;\;\; \tfscalk = \gamma^{-1}(s_k),
 \end{equation}
 where $s(t) = \gamma(t)$ is the particular timing law imposed during the trajectory demonstration, 
 describing how the robot moves along the imposed geometric path being variable $s$  the arc-length parameterization of the curve. 
From \eqref{new_eqq_2}, \eqref{norm_Delta1} and \eqref{Spatial_sequence}, the resulting trajectory parameterized with respect to the 
arc-length parameter $\hat\y_r(s)\approx \hat\y_L(s) = \yl(\gamma^{-1}(s))$ can be obtained. The 
 sequence $\y_{\Delta,k}$ can be viewed as the result of a sampling operation with a constant spatial period $\Delta$, characterized by:
\begin{equation}\label{unitary_norm}
{\left\lVert\dfrac{d \hat \y_r(s)}{d s}\right\rVert}_{s=s_k} \approx
\dfrac{\|\y_{\Delta,k+1}-\y_{\Delta,k}\|}{\|s_{k+1}-s_{k}\|}
=\!\dfrac{\Delta}{\Delta}\!
=1.
\end{equation}
%
%
{\black \begin{Remar}\label{curve_regular_remark}
The Spatial Sampling algorithm generates a filtered curve $\y_r(s)$ satisfying condition \eqref{unitary_norm}. From the latter, it follows that the filtered curve $\y_r(s)$ always satisfies the regularity condition~\eqref{regul_eq}, enabling the parameterization \eqref{eq16}
%
%
to be always well-defined.
\end{Remar}} 
The operation of the algorithm is graphically depicted in Fig.~\ref{SpSa_1D_example_05_1} for a one-dimensional case study.
Initially, by sampling the reference trajectory $\ylscal(\tr)$ at a constant time interval $T$, the linear interpolation function $\ylscal(t)$ is derived in Fig.~\ref{SpSa_1D_example_05_1}a. Note that the geometric distance of the samples $\ylscal(\tr)$
is clearly not constant. Subsequently, in Fig.~\ref{SpSa_1D_example_05_1}b and Fig.~\ref{SpSa_1D_example_05_1}c, the spatial sampling algorithm is applied to the
curve $\ylscal(t)$ using two different spatial intervals $\Delta$, where it can be noticed that smaller values of $\Delta$ result in a more accurate approximation of the original curve. {\black  This observation suggests considering the value of $\Delta$ as small as possible to maintain a meaningful representation of the demonstrated trajectory. However, Since $\Delta$ represents the amplitude of the minimum movement that can be captured by the curve, in applications involving humans -typically characterized by unwanted motions, such as tremors, superimposed on the desired demonstration- the choice of $\Delta$ can be used to filter out these undesired contributions. This can be achieved by setting the value of $\Delta$ larger than the magnitude of the maximum disturbance affecting the demonstrated trajectory. Note that this threshold can be easily estimated by asking the user to keep the robot's end-effector in its initial location for a given amount of time and measuring the maximum \textit{apparent} displacement.} \\
In Fig.~\ref{SpSa_1D_example_05_1}d, the samples of the timing law $s = \gamma(t)$ imposed during the demonstration of the trajectory and obtained with the same spatial period $\Delta$ of case (c)  are also presented. These illustrate how the length along the geometric path changes with time throughout the demonstration.  Fig.~\ref{SpSa_1D_example_05_1}a shows that the proposed spatial sampling imposes no constraints on the demonstrated trajectory $\y_r(t)$, which can also include parts with zero speed related to the user stopping during the execution. This is an important feature since no segmentation of the demonstrated trajectory is required, as it is by the arc-length (AL) DMP in~\cite{GASPAR2018225} instead. 
The pseudo-code of the Spatial Sampling algorithm is shown in Algorithm~\ref{algorithm:SS}\footnote{{\black All codes can be found at the online repository: \textit{https://github.com/AutoLabModena/Geometric-Dynamic-Movement-Primitives.git} }}.

 \begin{Remar}\label{GMDPdefinition}
The diagram depicted in Fig.~\ref{fig:DMPnewScheme}, where the parametric function $\y^\star(s)$ is computed based on the pair $(\sfk,\ \y_{\Delta,k})$, for $k=0,\ldots,M$, defines the concept of \textit{Geometric} DMP (GDMP). 
{\black From \eqref{unitary_norm} and \eqref{eq16}}, it follows that this type of DMP solely derives from the geometric path of the demonstrated trajectory and can be linked to any phase variable $s(t)$, {\black which can be freely chosen in order to generate a uniform velocity profile, thus improving the human experience.}
\end{Remar}

From Remark~\ref{GMDPdefinition}, the phase variable $s(t)$ can be chosen based on the three distinct scenarios depicted in Fig.~\ref{schematico}:

\noindent 1) Original: the demonstrated trajectory is reproduced using the original timing law $s(t) = \gamma(t)$.
    
\noindent 2) Offline Optimization: the phase variable $s(t)$ results from an optimization problem, which optimizes the motion along the geometric path to achieve a desired objective, as discussed in Sec.~\ref{Phase_Opt_DMP_section}. 


\noindent 3) Dynamical System: 
the phase variable $s(t)$ is defined by a dynamical system, possibly dependent on external inputs. This scenario includes co-manipulations tasks, as discussed in Sec.~\ref{human_in_the_loop_section} and Sec.~\ref{Experimental_Tests_section}.

\begin{figure}
\psfrag{x}[c][t][1]{$x$ [m] }
\psfrag{y}[c][t][1]{$y$ [m] }
\psfrag{z}[c][t][1]{$z$ [m] }
    \centering
    \includegraphics[width=0.9\columnwidth]{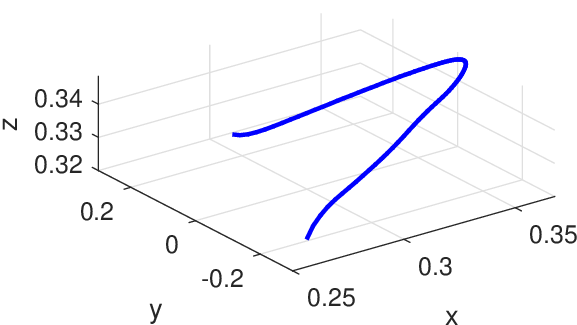}
    \caption{Demonstrated curve used for optimization.}
    \label{path4opti}
\end{figure}
\begin{figure}
\psfrag{t}[t][t][1]{$t$ [s]}
\psfrag{s}[c][t][1]{$s(t)$ [m] }
\psfrag{s1}[t][t][0.8]{$\gamma$}
\psfrag{s2}[t][t][0.85]{$\gamma_q$}
\psfrag{s3}[t][t][0.85]{$\gamma_s$}
\psfrag{s4}[t][t][0.85]{$\gamma_{qs}$}
    \centering
    \includegraphics[width=0.9\columnwidth]{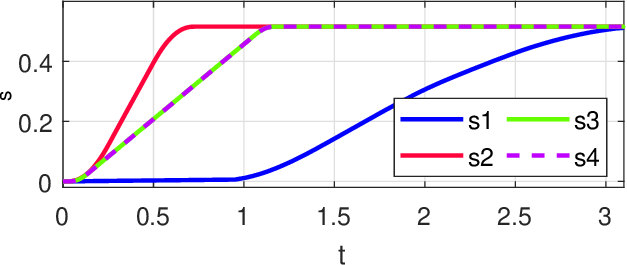}
    \caption{Phase profiles obtained as a solution of the optimization problem \eqref{opti_problemmm} subject to constraints (\ref{sder_conditions}) and (\ref{qder_conditions}).}
    \label{Optimal_S}
\end{figure}
\begin{figure}[t]
%
\psfrag{ds}[c][t][1]{$\dot{\overline{\gamma}}_s$} 
\psfrag{dds}[c][t][1]{$\ddot{\overline{\gamma}}_s$} 
\psfrag{dq normalized}[c][t][1]{$\dot{\overline{\q}}$} 
\psfrag{ddq normalized}[c][t][1]{$\ddot{\overline{\q}}$} 
\psfrag{t}[t][t][1]{$t$ [s]}
\psfrag{q1}[t][t][0.9]{$q_{1}$}
\psfrag{q2}[t][t][0.9]{$q_{2}$}
\psfrag{q3}[t][t][0.9]{$q_{3}$}
\psfrag{q4}[t][t][0.9]{$q_{4}$}
\psfrag{q5}[t][t][0.9]{$q_{5}$}
\psfrag{q6}[t][t][0.9]{$q_{6}$}
\psfrag{q7}[t][t][0.9]{$q_{7}$}

\centering
\begin{subfigure}{0.9\linewidth}
\centering
    \includegraphics[trim=0cm 0cm 0cm 0.015cm,clip=true,width=\columnwidth]{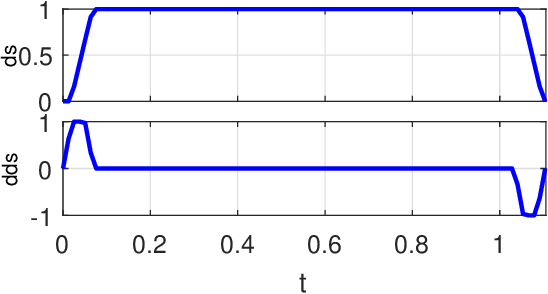}
    \caption{}
    \label{Optimal_WS}
\end{subfigure}\\[1mm]

\begin{subfigure}{0.9\linewidth}
\centering
    \includegraphics[trim=0cm 0cm 0cm 0.015cm,clip=true,width=\columnwidth]{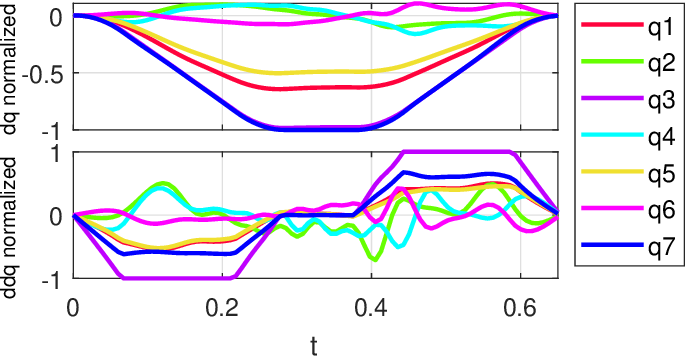}
        \caption{}
        \label{Optimal_Q}
\end{subfigure}
\caption{Normalized results of the optimization problem \eqref{opti_problemmm}
subject to (a) constraints~\eqref{sder_conditions} and (b) constraints~\eqref{qder_conditions}.} 
\label{fig:optimization_experiments}
\end{figure}

\section{{\black Phase Optimization in Geometric DMP}}\label{Phase_Opt_DMP_section}

Many planning techniques focus on minimizing 
the execution time while satisfying task and kinematic constraints~\cite{zoe2,vpsto,37,verscheure2009time,gattringer2022time}.
This optimization problem can be effectively addressed using the proposed GDMP framework, where the phase variable $s(t)$ is computed as the output of this dynamic system (i.e. a chain of two integrators):
\begin{equation}
    \dot{\x} = \left[ \begin{array}{ccc}
         0 & 1 & 0 \\
         0 & 0 & 1 \\
         0 & 0 & 0
    \end{array} \right] \x \\
    + \left[\begin{array}{c}
         0  \\
         0  \\
         1
    \end{array} \right] u,
\nonumber
\end{equation}
where $\x = \left[ s(t),\, \dot{s}(t),\, \ddot{s}(t) \right]^T $ and $u = \dddot{s}(t)$.
The optimization problem can be formulated as follows~\cite{gattringer2022time}:
\begin{equation}
\min_{u(\cdot)} \quad  T_f = \int_0^{T_f} 1 dt  = \int_0^{\Delta M} \frac{1}{\dot{s}(t)} ds(t),
\label{opti_problemmm}
\end{equation}
subject to:\\[-5mm]
 \begin{align}
  & s(0)=0, \quad s(T_f)=\Delta M,\nonumber \\
  & \dot{s}(0) = \dot{s}(T_f) = 0, \quad \dot{s} \geq 0, \nonumber\\
  & \ddot{s}(0) = \ddot{s}(T_f) = 0. \nonumber
 \end{align}
Additional constraints can be imposed on $\x$ and $u$:
\begin{equation}
        |\,\dot s\,(t)\,| \!\leq\!\dot{s}_{\MAX},  \;\;\;\;
        |\,\ddot s\,(t)\,| \!\leq\! \ddot{s}_{\MAX}, \;\;\;\;
        |\dddot s(t)| \!\leq\!\dddot{s}_{\MAX} ,
\label{sder_conditions}
\end{equation}
or on the robot velocity and acceleration in the task space:
\begin{equation}
\begin{array}{lllclll}
      \|\eta\, \dyp(s(t))\| &\!\!\!\!\leq\!\!\!\!& \dot{\y}_{\MAX} &\!\!\Leftrightarrow \!\!&       \|\, \dyp(s(t))\| &\!\!\!\!\leq\!\!\!\!& \dot{\y}_{\MAX}/\eta,  \\[1mm]
      \|\eta\, \ddyp(s(t))\| &\!\!\!\!\leq\!\!\!\!& \ddot{\y}_{\MAX} &\!\!\Leftrightarrow\!\!&       \|\, \ddyp(s(t))\| &\!\!\!\!\leq\!\!\!\!& \ddot{\y}_{\MAX}/\eta.
\end{array}
\label{yder_conditions}
\end{equation}
%
%
Alternatively, or in addition, constraints in the joint space can be considered as well, typically caused by the physical limitations of the actuation systems. Let $\qp(s(t))\in\mathbb{R}^{n}$, with $n$ the number of joints, be 
the joint trajectory profiles computed as follows:
{\black
\[
\qp (s(t))=\mbox{IK}\Big(\yp_\eta(s(t)), \R^\star(s(t))\Big),
\]
where IK$(\cdot)$ is the Inverse Kinematics function of the robot. Function $\yp_\eta(s(t))$ is the demonstrated (position) trajectory defined in \eqref{eq:y_eta}, while $\R^\star(s(t))$ is the desired orientation (in the simulation, $\R^\star(s(t))$ is kept constant but, in principle, can be also varied during the demonstration). } Velocity and acceleration constraints\footnote{\black The practical definition of the constraints \eqref{qder_conditions} is obtained by computing the joint configurations $\q_k$  from $(\eta \y_{\Delta,k}$, $\R_k)$, for $k=0,\ldots,M$,    using the IK algorithm. Then, the parametric  function $\qp (s)$ is deduced by applying the same algorithm used for $y^{\star}(s)$ in \eqref{eq8}. Finally, the constraints are defined as a function of the state $\x = \left[ s(t),\, \dot{s}(t),\, \ddot{s}(t) \right]^T $ and of ${\qp}' (s)$, ${\qp}'' (s)$ which are computed analytically from $\qp (s)$. Therefore, the two inequalities become 
\begin{equation}
\begin{array}{rrcll}
  -\dot{\q}_{\MAX}  &\!\leq\!&    {\qp}'(s(t))\dot{s}(t)  &\!\leq\!& \dot{\q}_{\MAX}, \\
  -\ddot{\q}_{\MAX}  &\!\leq\!&        {\qp}'' (s(t))\dot{s}^2(t) \!\!+\!{\qp}' (s(t))\ddot{s}(t) &\!\leq\!& \ddot{\q}_{\MAX}.
\end{array}
\nonumber
\end{equation}} in the joint space can be expressed as:
\begin{equation}
        |\dqp(s(t))|  \!\leq\! \dot{\q}_{\MAX}, \hspace{8mm}
        |\ddqp(s(t))| \!\leq\! \ddot{\q}_{\MAX}.
\label{qder_conditions}
\end{equation}
%
The considered optimization problem has been implemented in MATLAB using the CasADi optimal solver~\cite{Andersson2019}. {\black The reference curve  $\yp_\eta(s)$ is depicted in Fig.~\ref{path4opti} and the robot considered in the simulation is the same used  in the experimental tests described in Sec. \ref{Experimental_Tests_section}, i.e.  a Franka Emika Panda robot. As a 7-DOF robot with a redundant kinematic structure, the inverse kinematics is not available analytically and must be computed numerically (in particular, we used the inverse kinematics solver from the Matlab Robotics System Toolbox).}
Three different scenarios are considered: a)
 the constraints~\eqref{sder_conditions}, generating the timing law $s(t)=\gamma_s(t)$;
 b) the constraints~\eqref{qder_conditions}, generating the timing law $s(t)=\gamma_q(t)$; 
c) both the constraints~\eqref{sder_conditions} and~\eqref{qder_conditions}, generating the timing law $s(t)=\gamma_{qs}(t)$. 
The obtained results are compared 
in Fig.~\ref{Optimal_S} with the original timing law $\gamma(t)$.
The figure clearly shows that the execution time is significantly reduced compared to 
the demonstrated trajectory: $\gamma_s(t)$ and $\gamma_{qs}(t)$ result in an execution time $T_f=1.1$ s, denoting a reduction of $64.5\%$ with respect to $\gamma(t)$, while $\gamma_q(t)$ results in an execution time of $T_f=0.6$s, denoting a reduction of $80.6\%$ with respect to $\gamma(t)$. Indeed, one can conclude that the constraints~\eqref{sder_conditions} are more restrictive than~\eqref{qder_conditions}.
%
%
\begin{figure}[h!]
\psfrag{t}[t][t][1]{$t$ [s]}
\psfrag{x}[t][t][1]{$x$ [m]}
\psfrag{y}[t][t][1]{$y$ [m]}
\psfrag{z}[t][t][1]{$z$ [m]}
\psfrag{g0}[t][t][1]{$\g_r$}
\psfrag{y0}[t][t][1]{$\!\!\!\y_r(0)$}
\psfrag{g}[t][t][1.2]{$z$}
\psfrag{h}[t][t][1.2]{$\dot z$}
\psfrag{i}[t][t][1.2]{$\ddot z$}
\psfrag{aaaaa}[][][0.8]{DMP}
\psfrag{bbbbb}[t][t][0.8]{TCDMP}
\psfrag{ccccc}[][][0.8]{DMP$^*_P$}
\psfrag{eeeee}[t][t][0.8]{RDMP}
\psfrag{hhhhh}[t][t][0.8]{GDMP}
\psfrag{ddd}[][][1]{Ref}
\psfrag{y0}[t][t][1]{$\y_0$}
\psfrag{g0}[t][t][1]{$\g$}
\centering
\begin{subfigure}{0.9\linewidth}
\centering
    \includegraphics[trim=0cm 0cm 0cm 0.015cm,clip=true,width=\columnwidth]{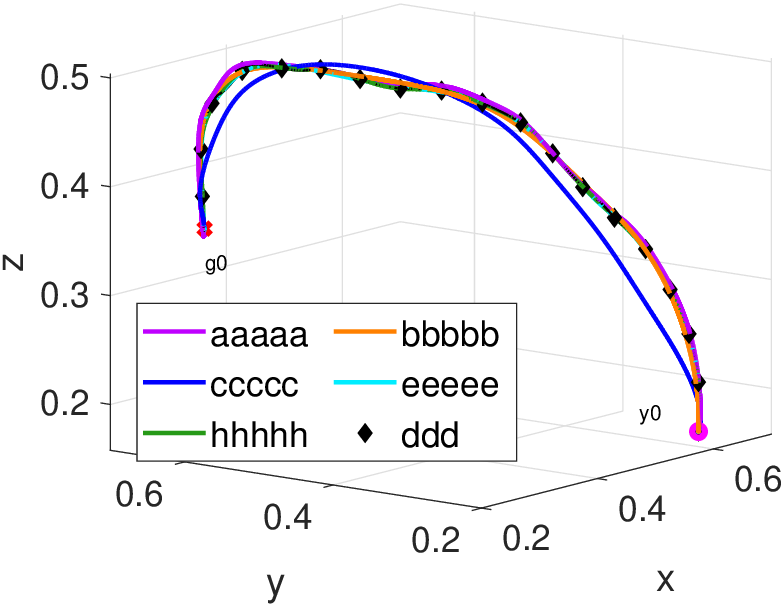}
    \caption{}
    \label{subfig:3d_path_opt}
\end{subfigure}\\[2mm]
\begin{subfigure}{0.9\linewidth}
\centering
    \includegraphics[trim=0cm 0cm 0cm 0.015cm,clip=true,width=\columnwidth]{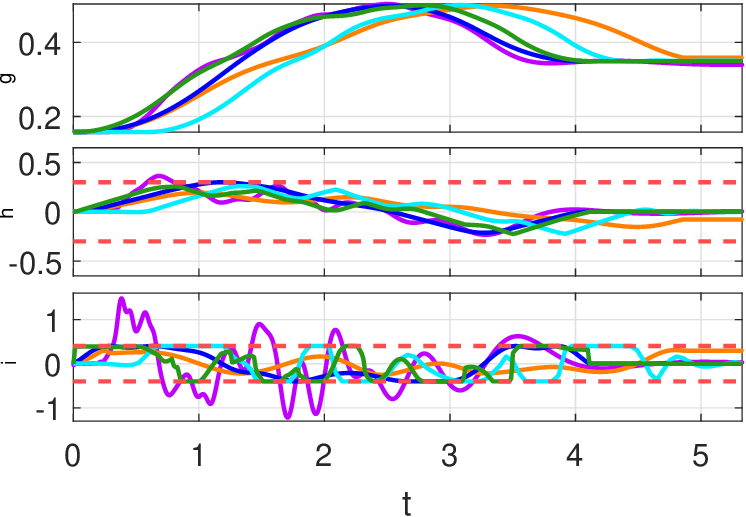}
    \caption{}
    \label{subfig:constr_opt}
\end{subfigure}\\[2mm]
\begin{subfigure}{0.9\linewidth}
    %
    \begin{center}
    \begin{tabular}{@{}c@{\,}c@{\;\;}c@{\;\;}c@{}}
    \hline
         & {\small Trajectory} & {\small Constraints} & 
         {\small Optimal} \\
         & {\small Reproduction} & {\small Satisfaction} & {\small Execution (min $T$)}\\
    \hline
         DMP & \gcheck & \rcross & \gcheck \\
         TCDMP & \gcheck & \gcheck & \rcross \\
         RDMP & \gcheck & \gcheck & \rcross\\
         DMP$^*_P$ & \rcross & \gcheck & \gcheck\\ 
         GDMP & \gcheck & \gcheck & \gcheck\\ 
    \hline
    \end{tabular}
        \end{center}
    \caption{}
    \label{subfig:table}
\end{subfigure}
\caption{ {\black Comparison of GDMP with other DMP solutions in addressing the minimum time optimization problem. (a) 3D path recordings; (b) $z$-axis' position, velocity and acceleration trajectories for each DMP model; (c) final evaluation table.} }
\label{fig:Opt_compare}
\end{figure}
The normalized profiles $\dot{\overline{\gamma}}_s=\dot\gamma_s(t)/\dot{s}_{\MAX}$ and $\ddot{\overline{\gamma}}_s=\ddot\gamma_s(t)/\ddot{s}_{\MAX}$ displayed in Fig.~\ref{Optimal_WS} confirm the optimality of the timing law $\gamma_s(t)$, since the boundary of at least one constraint in \eqref{sder_conditions} is reached at every time instant. The same conclusion can be drawn for the timing law $\gamma_q(t)$ by referring to Fig.~\ref{Optimal_Q}, presenting the normalized 
profiles $\dot{\overline{\q}}=
\dot{\q}(\gamma_q(t))/ \dot{\q}_{\MAX}$ and
$\ddot{\overline{\q}}=\ddot{\q}(\gamma_q(t))/ \ddot{\q}_{\MAX}$. In this case, the limiting factors are the velocity of joint $\#7$ and the acceleration of joint $\#3$.

\subsection{Comparison with other DMPs}\label{Comp_with_oth_DMPs_section}

{\black To demonstrate the effectiveness of the GDMP framework shown in Fig.\ref{schematico}, we applied it to solve the optimization problem~\eqref{opti_problemmm}, focusing on the task-space constraints described in~\eqref{yder_conditions}. To provide a comprehensive evaluation, we compared its performance with several approaches from the literature, including the classical DMP~\cite{5}, Temporal Coupling DMP (TCDMP)\cite{dahlin2021temporal}, the reversible DMP (RDMP)~\cite{zoe1}, and offline optimal DMP (DMP$^*_P$)\cite{zoe2}. These methods were applied to the same demonstrated trajectory, depicted in black in Fig.~\ref{subfig:3d_path_opt}. }

{\black For what concerns GDMP, the trajectory was processed, and the optimization problem~\eqref{opti_problemmm} was solved under the kinematic constraints~\eqref{sder_conditions} and~\eqref{yder_conditions}. The optimization yielded an optimal task duration of $T_f^\star =4.11s $. For consistency, the durations of the other methods were scaled to match $T_f^\star$, enabling a direct comparison of their adherence to the constraints and trajectory approximation. }

{\black The classical DMP achieves time scaling by modifying the temporal parameter $\tau$, preserving the original curve shape, as shown in purple in Fig.~\ref{subfig:3d_path_opt}. However, this approach does not inherently account for velocity or acceleration constraints, leading to violations of these constraints even when the desired task duration is achieved, see Fig.~\ref{subfig:constr_opt}. In contrast, TCDMP leverages temporal coupling to enforce kinematic constraints by appropriately adjusting $\tau$. While this ensures that the resulting trajectory stays within the constraint boundaries, it fails to meet the optimal duration $T^\star$, as the coupling mechanism indirectly affects task timing. Similarly, the RDMP structure allows addressing the optimization problem~\eqref{opti_problemmm} via $s^\star(t)$ but fails to achieve $T_f^\star = 4.11s$ due to its reliance on time-parametrized demonstrations, which encode speed variations and pauses. }

{\black Finally, DMP$^*_P$ addresses these limitations by directly optimizing the weights $\boldsymbol{\omega}$ in~\eqref{eq9}, ensuring both compliance with the constraints and the desired task duration. However, strict kinematic constraints can lead to a compromise in the curve approximation, as the optimization prioritizes constraint satisfaction over precise interpolation of the demonstrated trajectory, which can be observed in Fig.~\ref{subfig:3d_path_opt}. }

{\black The limitations of the existing approaches are summarized in the table of Fig.~\ref{subfig:table}, where the terms ``Trajectory Reproduction'', ``Constraints Satisfaction'', and ``Optimal Execution (min $T$)'' indicate the ability to reproduce the demonstrated trajectory, satisfy kinematic constraints, and achieve the optimal duration $T^\star$, respectively. While each existing DMP solution has specific drawbacks, the proposed GDMP, as highlighted in green in Fig.~\ref{subfig:table}, successfully fulfills all task requirements. }

{\black This success stems from the arc-length parametrization provided by the SS algorithm, which decouples the trajectory’s path and velocity information. This decoupling allows the velocity profile to be computed independently, ensuring compliance with kinematic constraints while maintaining the desired trajectory shape and task duration. Consequently, GDMP offers a robust solution for tasks requiring high precision, constraint adherence, and time optimization.
}

\begin{figure*}[t]
    \centering
    \includegraphics[width=1.86\columnwidth]{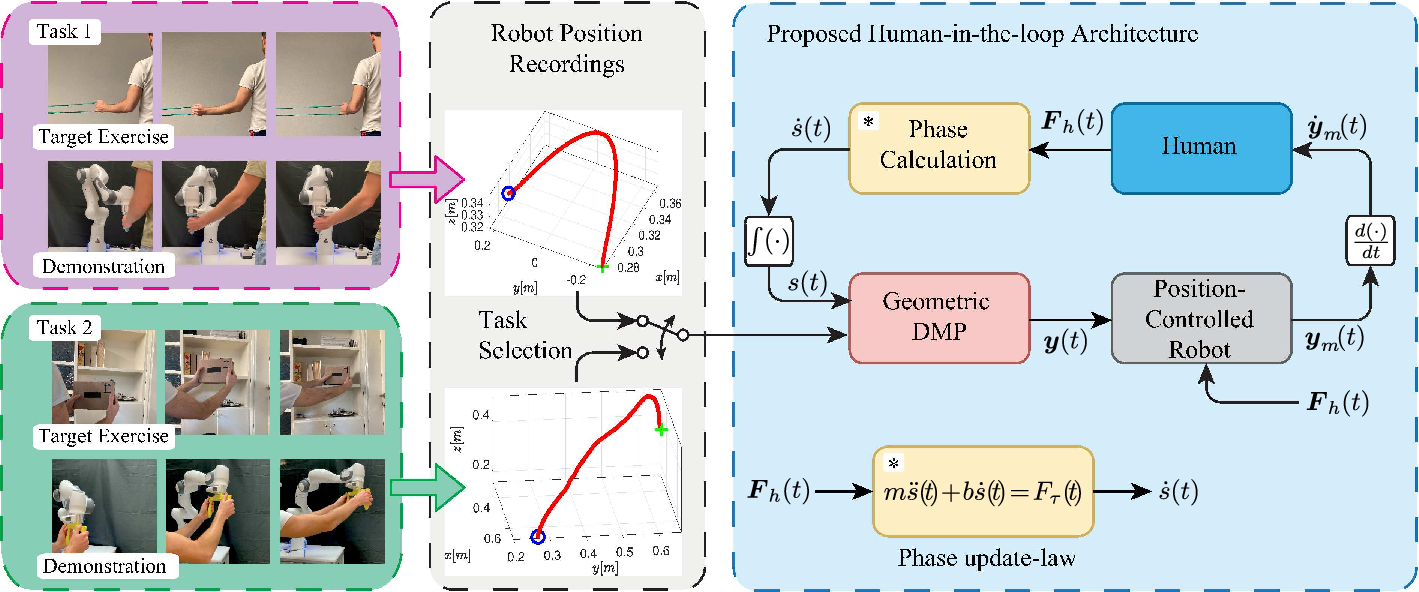}
\caption{{\black Human-robot interaction through kinesthetic guidance along the trajectories encoded by GDMP.}}\label{applic_1}
\end{figure*}

\section{Human-in-the-loop Geometric DMP}\label{human_in_the_loop_section}


The considered human-robot interaction framework
employing GDMP is schematically depicted in Fig.~\ref{applic_1}. The phase variable $s(t)$ is determined as

\begin{equation}\label{F_h_eq}
m\,\ddot{s}(t) + b\,{\dot{s}(t)} =F_{\tau}(t)= \y^{\star\prime}_\eta(s(t))^T\, \F_h(t),
\end{equation}
where $\y^{\star}_\eta(s)=\E \y^{\star}(s(t))$ is a generalization of \eqref{eq:y_eta} to the multidimensional case.
Equation~\eqref{F_h_eq} represents the dynamics of a mass $m$ with damping coefficient $b$, influenced by the external force $\F_h(t)$ applied to the robot end-effector by a human operator.
The presence of the scaling factor $\E$ in (\ref{F_h_eq}) brings the following two features: 

\noindent 1) It enables a consistent human-robot interaction,
as further discussed in the remainder of this section;

\noindent 2) It ensures the passivity of the human-robot interaction framework in Fig.~\ref{applic_1}, as proven in Sec.~\ref{Passivity_Analysis_section}. 


The force $F_{\tau}(t)$ in \eqref{F_h_eq}
%
%
denotes the projection of the human force vector $\F_h(t)$ onto the tangential direction of the reference trajectory $\y^{\star}_\eta(s)=\E \y^{\star}(s(t))$, where $\y^{\star}(s(t))$ is the parameterized trajectory as depicted in the GDMP schematic of Fig.~\ref{schematico}.
%
%
From \eqref{unitary_norm}, it follows that
$\|\y^{\star\prime}_\eta(s(t))\|\neq 0$, meaning that the projection in \eqref{F_h_eq} can always be successfully computed~\cite{trajpl,34,35}.
This presents an alternative solution to the approach proposed in~\cite{27}, where a minimum threshold is employed instead.

With reference to the aforementioned feature 1), the  
%
%
inclusion of matrix $\E$ ensures that the tangential direction remains consistent with the current orientation of the new curve. This feature proves valuable as it guarantees an accurate projection of the force $\F_h(t)$ in \eqref{F_h_eq} without the necessity of a new DMP parameterization~\cite{zoe3,zoe4}.

As shown in the framework of Fig.~\ref{applic_1}, an additional term $\F_h(t)$ is inserted in Eqs.~\eqref{eq13} and~\eqref{DMP_m_star}, leading to the following DMP formulation:
%
%
\begin{equation}\label{comb_DMP}
\begin{array}{@{\!\!\!\!\!}c}
\ddy(t) +\A \dy(t) + \A\B \y(t) -\A \B \g = \\[3mm]
\E \big[ \ddyp\!(s(t)) \!+ \!\!\A\dyp(s(t)) \!+\! \A\B \yp(s(t)) \!-\! \A\B \g_r \big] \!+\!\F_h(t),
\end{array}
\end{equation}
where  the additional term $\F_h(t)$ ensures the passivity of the human-robot interaction framework
of Fig.~\ref{applic_1}, as proven in Sec.~\ref{Passivity_Analysis_section}.
The motion trajectory $\y(t)$ is then fed into a robot manipulator equipped with high-gain position control, which
%
%
essentially implements an admittance control system for the physical human-robot interaction \cite{Keemink2018}. In this setup, the admittance model restricts the user's movement to follow the trajectories defined by the GDMP, allowing forward or backward motion. 
As shown in \cite{Keemink2018}, 
a feasible approach for ensuring a safe
human-robot interaction is to impose the passivity of the system with respect to the energetic port $\langle \dot{\y}_m(t), \,\F_h(t)\rangle$ describing the interaction between the human and the controlled robot. In case the human behaves passively, this condition ensures the stability of the system composed of the human and the robot. 



%
\subsection{Passivity Analysis}\label{Passivity_Analysis_section}

\begin{Definizione}\label{Def_1}
Let $P_T(t)=\sum_{i=1}^n P_i(t)=\u^T(t) \z(t)$ be the total
power flowing through the $n$ energetic ports of a dynamical system, where $\u(t)$ and $\z(t)$ are the input and output
vectors \cite{tebaldi2023unified}. Such a system is said to
be passive if there exists a non-negative storage function
$S(\x(t))$ of the state vector $\x(t)$ such that, for any
time interval $[t_0,\,t_1]$ and for any initial state vector
$\x_0$, and being $\x_1$ the state vector at time $t_1$, the following inequality holds true~\cite{khalil2002nonlinear,colgate1989interaction}:
{\black
\begin{equation}\label{passiv1}
S(\x_1)-S(\x_0) \le \int_{t_0}^{t_1} P_T(t)
\,dt \;
\leftrightarrow \;
\dot S(x(t)) \le  P_T(t).
\end{equation}
}
%

\end{Definizione}
\begin{Prop}\label{Prop_1}
The human-robot interaction framework of
Fig.~\ref{applic_1}
is passive with respect to the energetic port $\langle \dot{\y}_m(t), \,\F_h(t)\rangle$,
where $\dy_m(t)$ is the 
manipulator velocity.
\end{Prop}
{\em Proof.}
%
%
%
Let variables $\yt$,
$\dyt$ and $\ddyt$ be defined as:
\begin{equation}\label{new_variables}
\begin{array}{c}
\yt(t)=\y(t)- \g
-\E
\yp(s(t))+\E\g_r, \\[3mm]
\dyt(t)= \dfrac{d\, \yt(t)}{dt}= \dy(t)- \E
\dyp(s(t)), \\[3mm]
\ddyt(t)= \dfrac{d\, \dyt(t)}{dt}= \ddy(t)- \E \ddyp(s(t)),
\end{array}
\end{equation}
\noindent where $\g$ and $\g_r$ are the
goals of the desired and parameterized demonstrated trajectories $\y(t)$ and $\yp(s(t))$, respectively. Using~\eqref{new_variables}, ~\eqref{comb_DMP} can be
rewritten as:
%
%
\begin{equation}\label{comb_DMP_1}
\ddyt(t)+\A \dyt(t)+\A\B
\yt(t)=\F_h(t),
\end{equation}
%
%
The following storage function
$S(\yt(t),\,\dyt(t))$ for the system
in \eqref{F_h_eq} and \eqref{comb_DMP_1} can be considered:
%
%
\begin{equation}\label{S_Delta}
\!\!\!\! S(\yt(t),\dyt(t))\!=\!
\underbrace{
\dfrac{1}{2}\dyt^T\!\!(t)
\dyt(t)}_{E_{K_1}}
\!+\!\underbrace{\dfrac{1}{2}\yt^T\!\!(t)
\A\!\B\yt(t)}_{E_{U}} \!+\!
\underbrace{\dfrac{1}{2}m\dot{s}^2(t)}_{E_{K_2}},
\end{equation}
{\black
where $E_{K_1}$ and $E_{U}$ represent the kinetic and potential energies, respectively, stored in the Geometric DMP 
%
%
of Fig.~\ref{applic_1}, while $E_{K_2}$ is the kinetic energy of the Phase Calculation block 
of Fig.~\ref{applic_1}.
}
%
%
Differentiating the storage function
$S(\yt(t),\,\dyt(t))$ in \eqref{S_Delta} with respect to time yields:
\begin{equation}\label{newwww_4}
\dot{S }(t) = \dyt^T\!\!(t) \big[ \ddyt(t) + \A\B \yt(t) \big] + m \dot{s}(t) \ddot{s}(t).
\end{equation}
Substituting $\ddyt(t)$ from \eqref{comb_DMP_1} in \eqref{newwww_4} yields:
%
\begin{equation}\label{newwww_5}
\begin{array}{@{\!\!}r@{}c@{}l}
\dot{S}(t) &=&\dyt^T\!(t) \big[ \!-\! \A \dyt(t) \!-\! \A \B \yt(t) \!+\! \F_h(t) \!+\! \A \B \yt(t) \big] + \\[2mm]
      &&+ m \dot{s}(t) \ddot{s}(t)\,.
\end{array}
\end{equation}
%
Substituting $\ddot{s}(t)$ from \eqref{F_h_eq} in \eqref{newwww_5} yields:
\begin{equation}\label{newwww_6}
\begin{array}{r@{\,}c@{\,}l}
     \dot{ S}(t) & = & - \dyt^T(t) \A \dyt(t) + \dyt^T(t)\F_h(t) + \\[2mm]
         && - b \dot{s}^2(t) + \dot{s}(t) \y^{\star\prime}(s(t))^T\!\E \F_h(t)\,.
\end{array}
\end{equation}
\begin{figure}[t]
    \centering
    \begin{subfigure}{0.15\textwidth}
        \includegraphics[width=\linewidth]{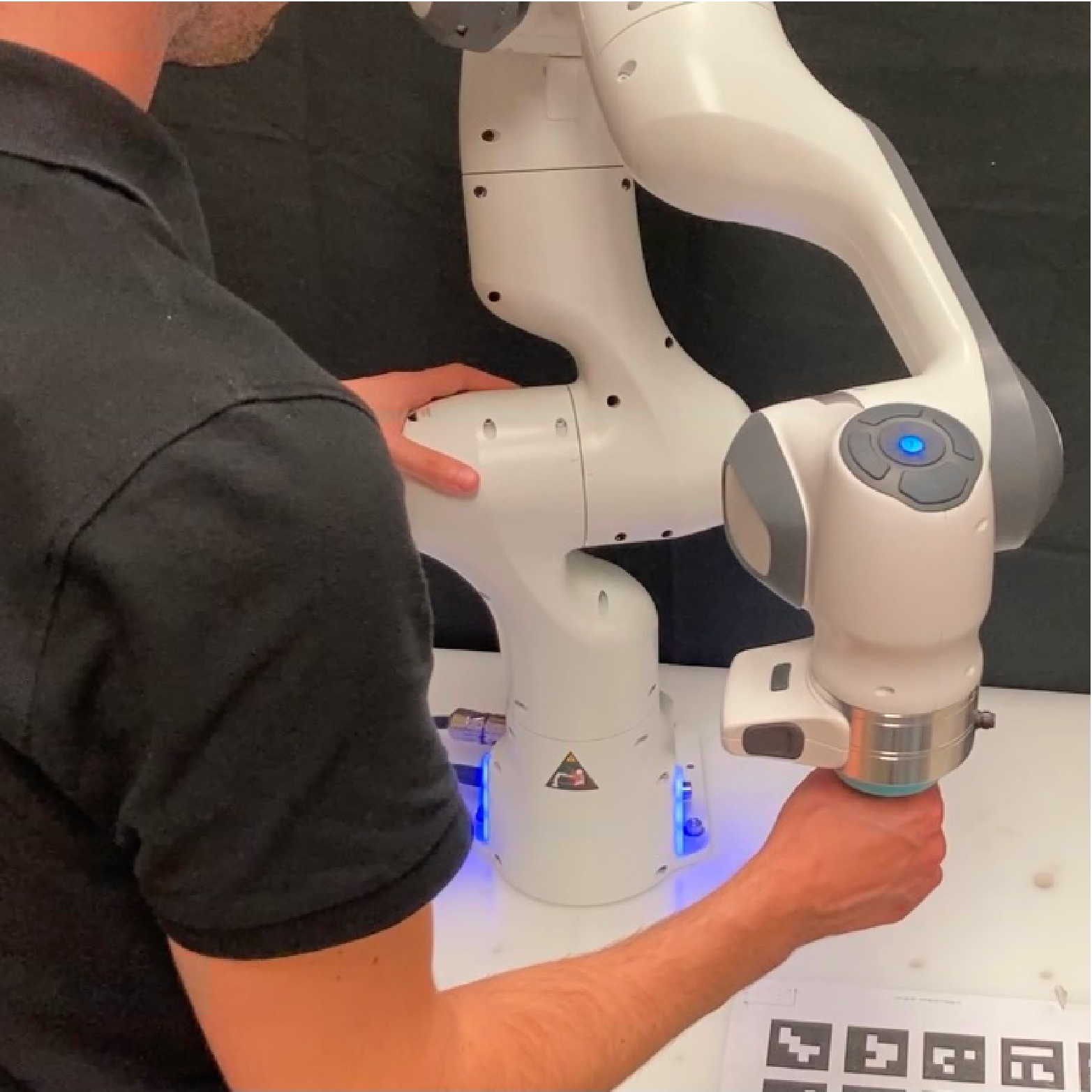}
        \caption{} \label{task1}
    \end{subfigure}%
    \hfill
    \begin{subfigure}{0.15\textwidth}
        \includegraphics[width=\linewidth]{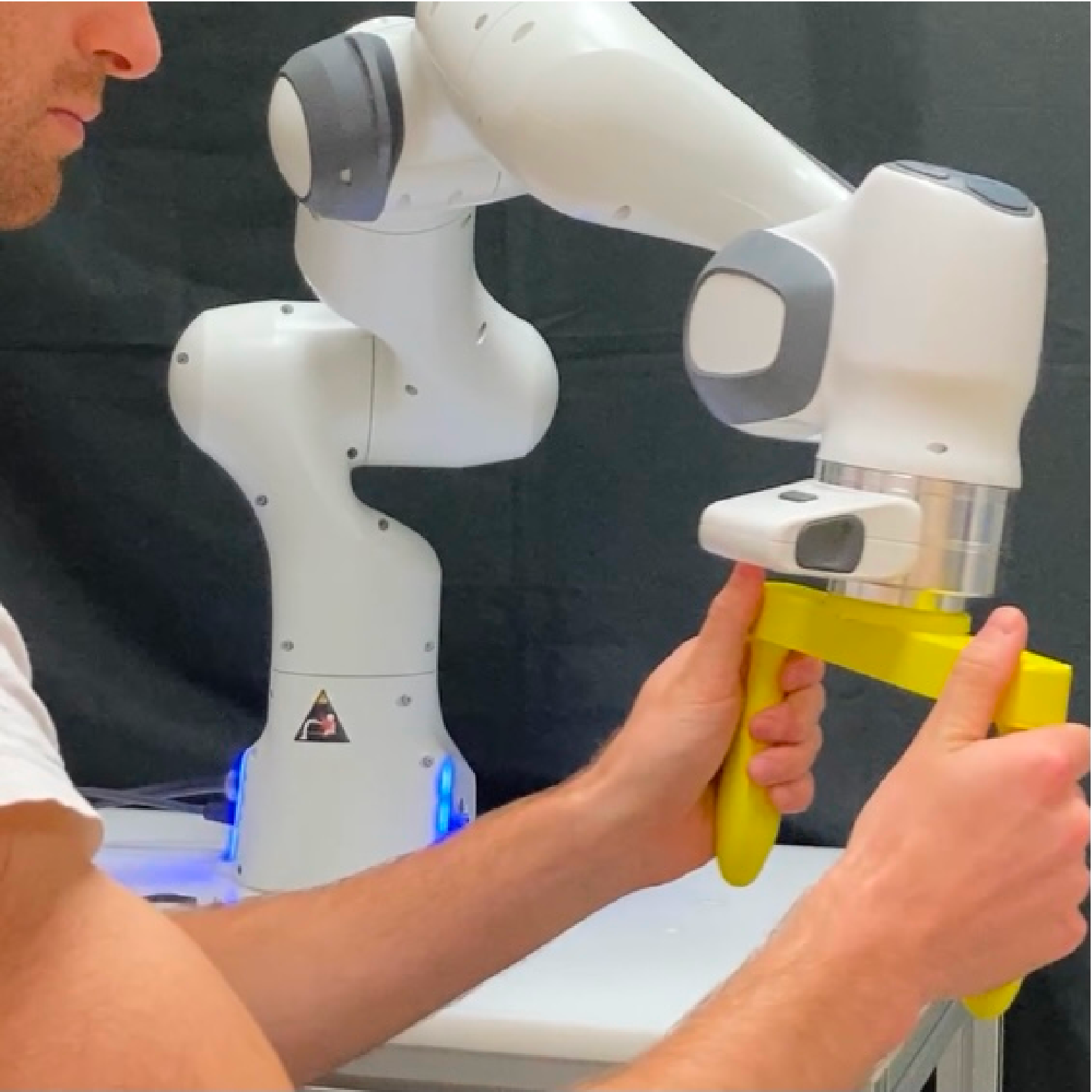}
        \caption{} \label{task2}
    \end{subfigure}%
    \hfill
    \begin{subfigure}{0.15\textwidth}
        \includegraphics[width=\linewidth]{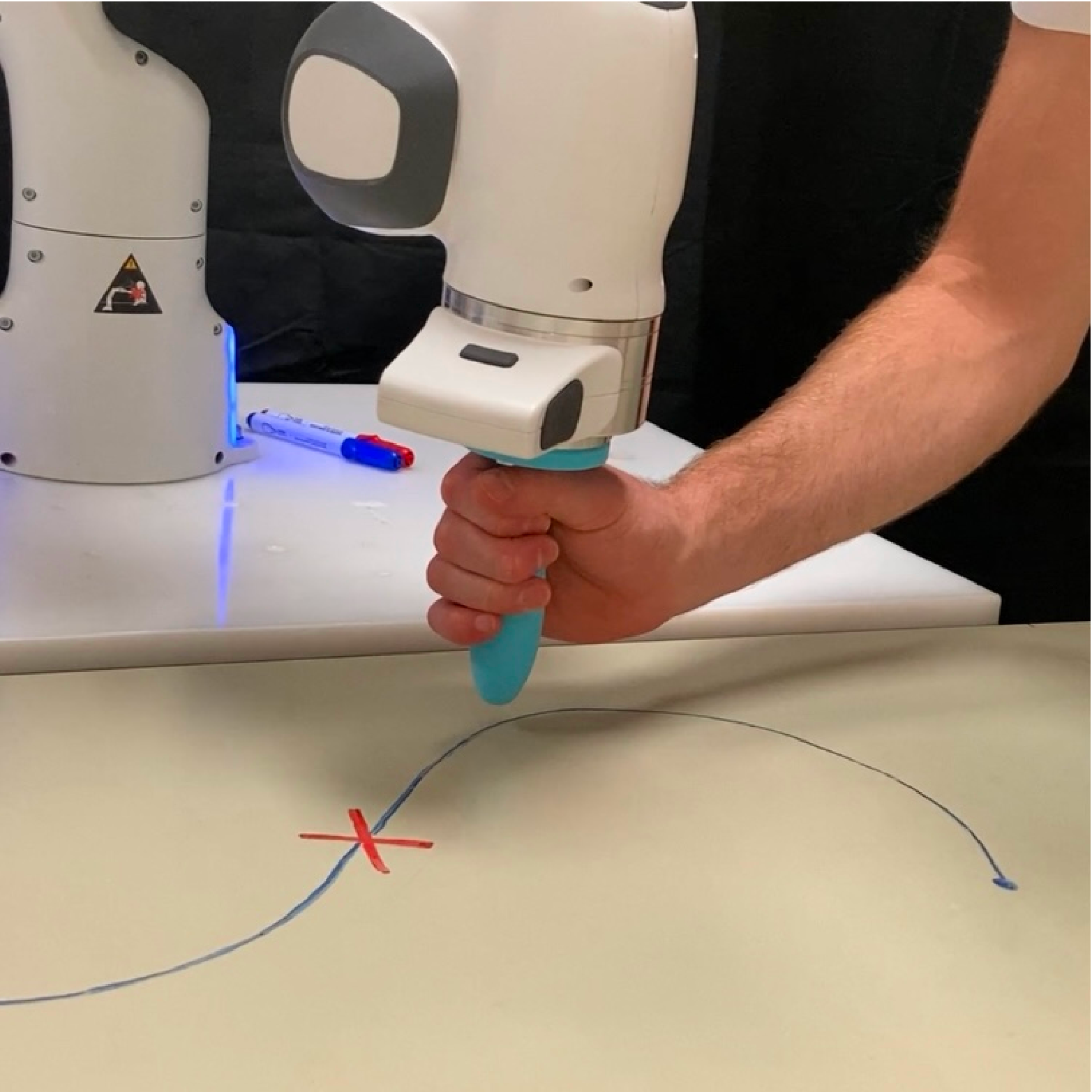}
        \caption{} \label{task3}
    \end{subfigure}%

    \caption{The considered experimental tests: (a) Internal/External rotation of the shoulder, (b) Bi-manual grabbing, (c) Stop at a point. }
    \label{tasks}
\end{figure}
Finally, substituting $\dyt(t)$ from \eqref{new_variables} in \eqref{newwww_6} and recalling that $ \y^{\star\prime}(s(t)) \dot{s}(t)  = \dyp(s(t)) $ from \eqref{eq16} yields:
\begin{equation}\label{S_Delta_1}
\begin{array}{r@{}c@{}l}
 \dot{ S}(t) & = & - \dyt^T\!(t) \A \dyt(t) \!+ \!\big[ \dy(t) - \E \dyp(s(t)) \big] ^T \F_h(t) + \\[3mm]
         && -b \dot{ s}^2(t) + \dy^{\star T}\!(s(t)) \E \F_h(t),
\\[3mm]
&  = &  -\dyt^T\!(t) \A \dyt(t) + \dy^T\!(t) \F_h(t) - b \dot{ s}^2(t).
\end{array}
\end{equation}
Considering the power $P_T=\dy_m^T\!(t) \F_h(t)$ flowing through the energetic port $\langle \dot{\y}_m(t), \,\F_h(t)\rangle$ and using \eqref{S_Delta_1}, the passivity condition \eqref{passiv1} is fulfilled. Indeed, assuming perfect tracking by the robot of the reference trajectory generated by the GDMP, i.e. $\y_m(t)=\y(t)$ and consequently $\dy_m(t)=\dy(t)$, the inequality
\begin{equation}\label{S_Delta_2}
    -\dyt^T\!(t) \A \dyt(t) + \dy^T\!(t) \F_h(t) - b \dot{s}^2(t) \le \dy_m^T\!(t) \F_h(t)
\end{equation}
{\black
transforms into $-\dyt^T\!(t) \A
\dyt(t)
-b \dot{s}^2(t) \le 0$,
%
%
%
which is always satisfied for $\A>0$ and $b>0$.
}
\section{{\black Human-Robot Interaction: Experimental Evaluation}}\label{Experimental_Tests_section}

This section presents and discusses the experimental tests conducted to validate the 
human-robot interaction framework of Fig.~\ref{applic_1}.
 Section~\ref{Experimental_Setup} outlines the utilized setup and details the
 {\black experimental tests performed for the tuning of parameters $m$ and $b$ in \eqref{F_h_eq} through the practical stability analysis of Sec.~\ref{Parameters_Tuning}. The results  have shown the emergence of instability phenomena such as vibrations and oscillations despite the passivity proof of Sec.~\ref{Passivity_Analysis_section}, where perfect tracking of the GDMP trajectory is assumed. This motivates the introduction of the stability analysis performed in Sec.~\ref{Parameters_Tuning}.
 Finally, Section~\ref{Insertion_task} addresses the comparison of GDMP with other DMP solutions with reference to an insertion task, showing the enhanced performance of GDMP\footnote{ {\black The video can be found at: \textit{https://youtu.be/cMcBOLuNb44} } } .
 }
 

\subsection{Experimental Setup}\label{Experimental_Setup}



The experimental setup involves the use of
a Franka Emika Panda robot in the GDMP framework of Fig.~\ref{applic_1} for co-manipulation tasks.
%
%
Three distinct experimental tests have been designed as illustrated in Fig.~\ref{tasks}, requiring the user to manipulate the robot end-effector along the desired position trajectory $\y(t)$ generated by the GDMP.
The manipulator position control is
developed in the MATLAB/Simulink environment. 
%
%
The experiments involved 12 participants of different gender, aged between 20 and 60, out of whom only three possessed prior expertise in robotics. All of whom provided informed consent prior to participation.

%
%
In the test of Fig.~\ref{task1}, the user is required to perform an internal and external rotation of the shoulder in the upright position. In the test of Fig.~\ref{task2}, the user is required to lead the end-effector from the initial to the goal position using a bi-manual grabbing in a seated position. Finally, in the experiment of Fig.~\ref{task3}, the user is required to move the end-effector from the initial position to a given point located along the demonstrated trajectory.
For each task, an expert user demonstrated the reference position trajectory $\yr(t)$ through kinesthetic guidance. The parameterized position trajectory $\yp(s(t))$ has then been generated as depicted in Fig.~\ref{schematico}.


\begin{figure}[t]
\psfrag{task1}[t][t][0.85]{\!\!task (a)}
\psfrag{task2}[t][t][0.85]{\!\!task (b)}
\psfrag{F}[t][t][0.85]{$\cos{(\theta)}_{RMS}$}
\psfrag{b}[t][t][0.85]{$b$}
\psfrag{m}[t][t][0.85]{$m$}
\psfrag{p}[t][t][0.85]{$\dot s_{RMSD}$}
\centering
\begin{subfigure}{0.9\linewidth}
\centering
    \includegraphics[trim=0cm 0cm 0cm 0.015cm,clip=true,width=\columnwidth]{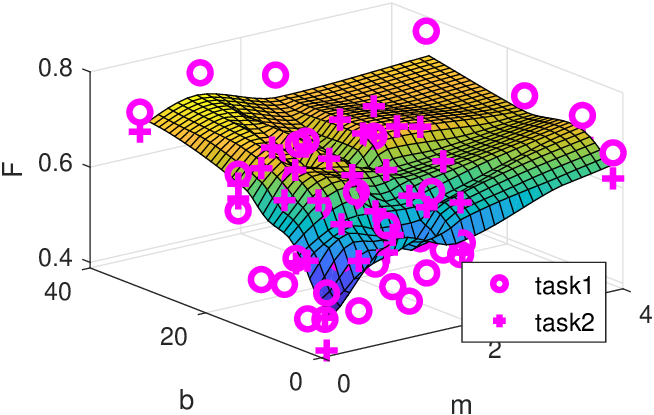}
    \caption{}
    \label{ft}
\end{subfigure}\\[1mm]

\begin{subfigure}{0.9\linewidth}
\centering
    \includegraphics[trim=0cm 0cm 0cm 0.015cm,clip=true,width=\columnwidth]{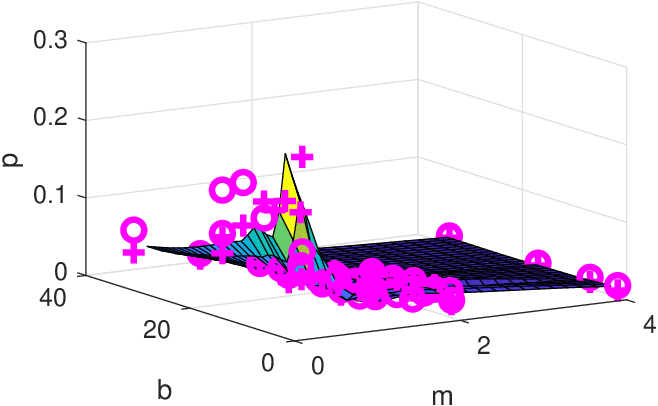}
        \caption{}
        \label{epp}
\end{subfigure}
\caption{Trend of a) external force application angle $\cos(\theta)_{RMS}$ and b) peak-to-peak oscillations $\dot s_{RMSD}$ on the phase velocity. }
\label{fig:ab_experiments}
\end{figure}
\begin{figure}[t]
\psfrag{t}[t][t][0.9]{$t$ [s]}
\psfrag{s}[t][t][1.]{\hspace{-2mm}$s(t)$}
\psfrag{P}[t][t][1.]{$P_\tau(t)$\hspace{0pt}}
\psfrag{m=0.2 b=7}[t][t][0.8]{$m\!=\!0.2, \ b\!=\!7$}
\psfrag{m=2 b=17}[t][t][0.8]{$m\!=\!2, \ b\!=\!17$}
    \centering
    \includegraphics[clip,width=0.9\columnwidth]{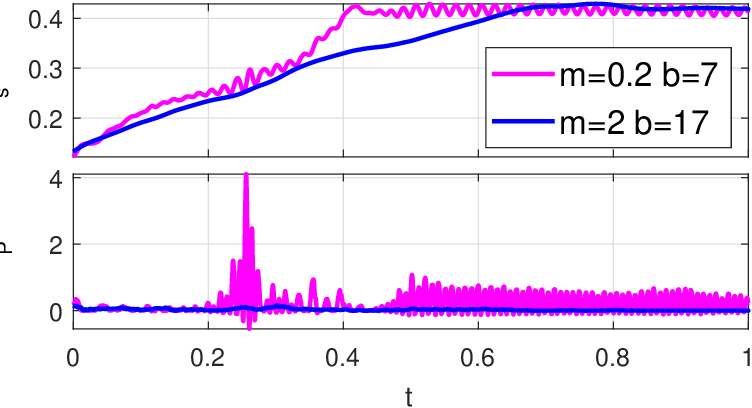}
  \setlength{\unitlength}{5.0mm}
 \psset{unit=\unitlength}
  \rput(-1.75,1.25){
  \rput(-12.555,6.68){\footnotesize (a)}
  \rput(-12.555,3.08){\footnotesize (b)}
  }
    \caption{(a) Phase variable $s(t)$ and (b) power $P_\tau(t) = F_\tau(t) \dot s(t)$ exchanged between human and the robot.    
    }
\label{compare_sp}
\end{figure}

{\black
\subsection{Parameters Tuning through Practical Stability Analysis}\label{Parameters_Tuning}
In the GDMP of ~\eqref{DMP_m_star}, 
%
%
the parameters $\alpha_x=\alpha_y=\alpha_z=40$ and $\beta_x=\beta_y=\beta_z=10$ have been assigned to the matrices $\A$ and $\B$, in order to ensure a sufficiently high stiffness 
in \eqref{comb_DMP_1} guaranteeing $\y(t) \approx \yp(s(t))$.
%
The choice of $m$ and $b$ in 
~\eqref{F_h_eq} is pivotal in practical applications. In order to extract
first-attempt values for $m$ and $b$, the Franka Emika Panda specifications have been considered:
%
%
%
$\dot{\y}_{\MAX} =  1.7 \ \textrm{m/s}$, $ \ddot{\y}_{\MAX}=13.0 \ \textrm{m/s}^2$ and a 
maximum payload of $3$ kg, which implies a force limit of $F_{\MAX}=30 N$. 
Under the nominal condition $\E=\I_3$ and recalling \eqref{unitary_norm},
it follows that $\|\dyp(s(t))\| = |\dot{s}(t)|$ and, if the curvature is negligible, $\|\ddyp(s(t))\| \approx |\ddot{s}(t)|$ from \eqref{eq16}. 
Consequently, the following specifications can be derived:
\begin{equation}
\label{eq:FrankLimits}
\dot{s}{\MAX} = 1.7 \ \textrm{m/s} \,\,\,\text{ and }\,\,\, \ddot{s}{\MAX} = 13.0 \textrm{m/s}^2.
\end{equation}
%
When the velocity $\dot{s}(t)$ is sufficiently low, the following values of values of $m$ and $b$ 
can be determined from ~\eqref{F_h_eq} and~\eqref{eq:FrankLimits}:
%
\begin{equation}\label{eq:m_b_values}
m = \frac{F_{\MAX}}{\ddot{s}{\MAX}} \approx 2 \ \textrm{kg} \quad \textrm{and} \quad b = \frac{F{\MAX}}{\dot{s}_{\MAX}} \approx 17 \ \textrm{N\,s/m}.
\nonumber
\end{equation}
Starting from~\eqref{eq:m_b_values},  
the following ranges have been experimentally investigated:
\begin{equation}\label{b_m_sets}
\cN_m=[0.2,\;\; 4] \;\mbox{kg} \,\,\,\text{ and }\,\,\,
\cN_b=[1.7,\;\; 34] \;\mbox{N$\cdot$ s/m}.
\end{equation}
  }
{\black
The three tests in Fig.~\ref{tasks} have been executed by each user for $60$ s using $m$ and $b$ in \eqref{b_m_sets}.
The results of the tests of Fig.~\ref{task1} and Fig.~\ref{task2} are shown in Fig.~\ref{fig:ab_experiments}.
The metric in Fig.~\ref{ft} is defined as follows:
\begin{equation}
\cos{\!(\theta)}_{RMS}\!=\!\mbox{rms}\big(\!\cos{({\theta}(t))}\big)\!, \hspace{1mm}\mbox{with}\hspace{1mm}
\cos{\!({\theta}(t))}\!=\!\dfrac{F_\tau(t)}{\|\F_h(t)\|},
\nonumber
\end{equation}
%
giving the normalized effort that the user has to make in order to move along the desired trajectory. 
The metric in Fig.~\ref{epp} is defined as follows:
\begin{equation}\label{s_dot_filt}
\nonumber
\dot{s}_{RMSD}=\text{rms}\big(\dot{s}(t)-\dot{s}(t){_\text{AVG}}\big),
\end{equation}
where $\dot{s}(t)_{\text{AVG}}$ is the moving average of the function $\dot{s}(t)$. 
This metric quantifies the average magnitude of the oscillations affecting the phase velocity $\dot{s}(t)$.

%

%
 The surface plot of Fig.~\ref{ft}
suggests that, as parameters $m$ and $b$ decrease, the user effort required to execute the task results to be lower as expected. However, as shown in Fig.~\ref{epp}, this also leads to higher oscillations affecting the phase velocity $\dot{s}(t)$, which may cause the robot to stop because of violation of the safe limit constraints. These oscillations are also evident in Fig.~\ref{compare_sp}, showing the results of the test in Fig.~\ref{task3}. Consequently, for lower values of $m$ and $b$ falling within the yellow peak in Fig.~\ref{epp}, 
the user's attempt to counteract the robot oscillations results in a larger recorded power $P_\tau(t)$.}


{\black
%
%

This oscillating behavior can be explained using the following practical stability analysis on the system of Fig.~\ref{applic_1} based on the Lyapunov's indirect method~\cite{khalil2002nonlinear}. Although the robot is typically assumed to perfectly track the desired reference, the bandwidth of the position control (Fig.~\ref{applic_1}) is finite in practical applications. To account for this, a finite delay $t_0$ between the robot position $\y_m(t)$ and the desired reference $\y(t)$ has been introduced.
Although providing an accurate description of the human is non-trivial~\cite{ficuciello2014cartesian}, the following simplified spring-damper model exhibiting isotropic behavior across all spatial directions is employed in this work:
%

\begin{equation} 
\nonumber
\begin{array}{c}
    \F_h
   =
    G_h(s_l)\,\I_3\,
    \dy_m 
    ,\hspace{5.6mm}  \mbox{where} \hspace{5.6mm}
    G_{h}(\!s_l\!)  =  \frac{B_{h}s_l+K_{h}}{s_l}
    \end{array}
\end{equation}
and where $s_l$ is the Laplace complex variable.
\begin{figure}[t]
\centering \small \setlength{\unitlength}{3.1mm}
\psset{unit=\unitlength} \SpecialCoor

\newrgbcolor{greenn}{0 0.5 0}
\newrgbcolor{viola_l}{0.8, 0.6, 1}
\newrgbcolor{viola}{0.4, 0.2, 0.6}
\newrgbcolor{bluD}{0.09, 0.45, 0.81}
\newrgbcolor{bluD_l}{0.29, 0.65, 1}
\newrgbcolor{rosso}{0.72, 0.34, 0.31}
\newrgbcolor{rosso_l}{0.97, 0.81, 0.8}
\newrgbcolor{giallo}{0.84, 0.71, 0.34}
\newrgbcolor{giallo_l}{1, 0.95, 0.8}
\newrgbcolor{grigio}{0.4, 0.4, 0.4}
\newrgbcolor{grigio_l}{0.96, 0.96, 0.96}
\begin{pspicture}(1,-3)(26,10)
\rput(0,-1){
\rput(1,0){
\psline[linearc=.2,linecolor=giallo,fillstyle=solid,fillcolor=giallo_l](5.5,0.5)(5.5,2.)(10.5,2.)(10.5,-1.)(5.5,-1.)(5.5,0.5)
\rput(8,0.5){\scriptsize $\dfrac{1}{(ms_l+b)}$}
}
\psline[linearc=.2,linecolor=black,linewidth=0.8pt]{->}(11.5,0.5)(15.5,0.5)
%
\rput(-1,0){
\psline[linearc=.2,linecolor=grigio,fillstyle=solid,fillcolor=grigio_l](16.5,0.5)(16.5,2)(21.5,2)(21.5,-1)(16.5,-1)(16.5,0.5)
\rput(19.001,0.5){\scriptsize $e^{-t_0s_l}$}
}
\psline[linearc=.2,linecolor=black,linewidth=0.8pt]{->}(20.5,0.5)(25,0.5)(25,3)
\psline[linearc=.2,linecolor=black,linewidth=0.8pt]{->}(25,5)(25,7.5)(23,7.5)
%
\rput(-1,1){
\psline[linearc=.2,linecolor=rosso,fillstyle=solid,fillcolor=rosso_l](19.,6.5)(19.,8.)(24.,8.)(24.,5.)(19.,5.)(19.,6.5)
\rput(21.5,6.5){\scriptsize $\dfrac{\partial \y^\star\!(s)\!}{\partial s}$}
}
\psline[linearc=.2,linecolor=black,linewidth=0.8pt]{->}(18.,7.5)(16,7.5)
%
\rput(0,1){
\rput(1,0){
\psline[linearc=.2,linecolor=giallo,fillstyle=solid,fillcolor=giallo_l](3.,6.5)(3.,5.)(8.,5.)(8.,8.)(3.,8.)(3.,6.5)
\rput(5.5,6.5){\scriptsize $\dfrac{\partial \y^\star\!(s)\!^T}{\partial s}$}
}
\psline[linearc=.2,linecolor=black,linewidth=0.8pt]{->}(11.,6.5)(9.,6.5)
\psline[linearc=.2,linecolor=bluD,fillstyle=solid,fillcolor=bluD_l](11.,6.5)(11.,5.)(16.,5.)(16.,8.)(11.,8.)(11.,6.5)
\rput(13.4,6.5){\scriptsize $\G_h(s_l)$}
}
\psline[linearc=.2,linecolor=black,linewidth=0.8pt]{->}(2,3)(2,0.5)(6.5,0.5)
\psline[linearc=.2,linecolor=black,linewidth=0.8pt]{->}(4.,7.5)(2,7.5)(2,5)
%
\psline[linearc=.2,linecolor=nero,fillstyle=solid,fillcolor=bianco](1,4)(1,3)(3,3)(3,5)(1,5)(1,4)
\rput(2,4){\scriptsize $\frac{1}{s_l}$}
%
\psline[linearc=.2,linecolor=nero,fillstyle=solid,fillcolor=bianco](24,4)(24,3)(26,3)(26,5)(24,5)(24,4)
\rput(25,4){\scriptsize $s_l$}
%
%
\psline[linearc=.2,linecolor=rosso, linestyle=dashed](3.5,7.5)(3.5,5.5)(23.5,5.5)(23.5,9.5)(3.5,9.5)(3.5,7.5)
\rput(13.4,4.5){ $G_h(s_l)$}
}
\end{pspicture}
\caption{{\black Linearized Human-robot interaction framework through kinesthetic guidance.}}\label{applic_2}
\vspace{2mm}
\end{figure}

\begin{figure}[t]
\psfrag{Re}[t][t][1]{Re}
\psfrag{Im}[t][t][1]{Im}
\psfrag{m1}[t][t][0.68]{$-1$}
\psfrag{t0}[t][t][0.68]{$t_0$}

\psfrag{a}[b][b][0.85]{(a)}
\psfrag{b}[b][b][0.85]{(b)}
    \centering
    \includegraphics[width=0.9\columnwidth]{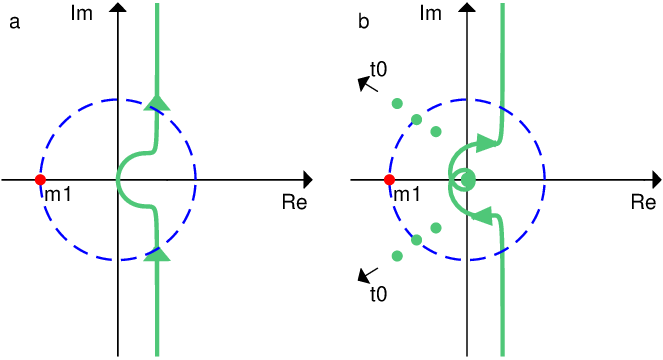}
    \caption{Nyquist diagram of the loop-gain function $G_{LG}(s_l)$ without (a) and with (b) finite delay $t_0$. }
    \label{nyquist_fig}
\end{figure}
\begin{figure}[t]
\psfrag{m}[t][t][0.9]{$m$}
\psfrag{b}[t][t][0.9]{$b$}
\psfrag{P}[t][t][0.8]{$M_\varphi$ [deg]}
    \centering
    \begin{subfigure}{0.15\textwidth}
        \includegraphics[width=\linewidth]{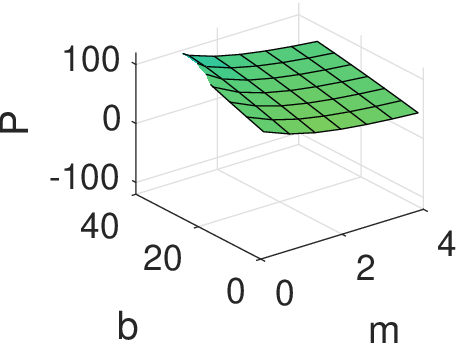}
        \caption{} \label{PM00_bis}
    \end{subfigure}%
   \hfill
    \begin{subfigure}{0.15\textwidth}
        \includegraphics[width=\linewidth]{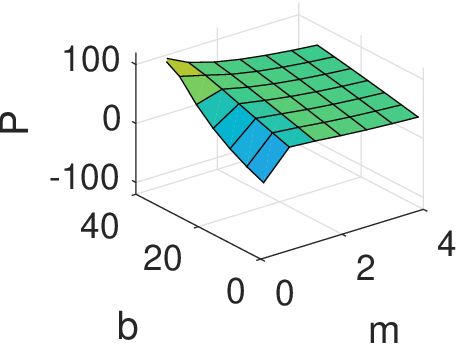}
        \caption{} \label{PM01_bis}
    \end{subfigure}%
   \hfill
    \begin{subfigure}{0.15\textwidth}
        \includegraphics[width=\linewidth]{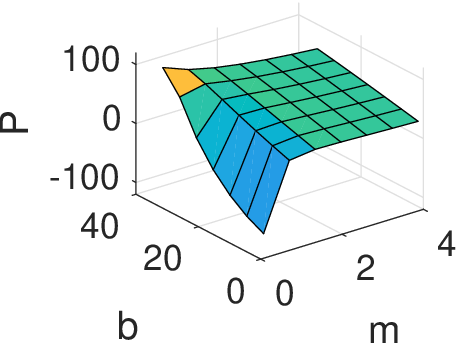}
        \caption{} \label{PM02_bis}
    \end{subfigure}%

    \caption{Phase margin $M_\varphi$  of the loop-gain function $G_{LG}(s_l)$ as a function of $m$, $b$ for different values of the delay $t_0$: $t_0\!=\!0$ s (a),  $t_0\!=\!0.01$ s (b), and $t_0\!=\!0.02$ s (c).} 
    \label{PhaseMargin_bis}

\end{figure}
Assuming that the GDMP behaves as an ideal trajectory generator, i.e. $\y(t) \approx \yp(s(t))$, and linearizing the system of Fig.~\ref{applic_1} 
around the equilibrium point $(s_e,\,\dot s_e) = (\bar s,\,0)$, where $\bar s$ represents the stop position, the scheme 
illustrated in Fig.~\ref{applic_2} can be derived, which
%
is characterized by the following input-output transfer function:
\begin{equation}\label{G_io_G_lg}
 \begin{array}{@{\!\!}cc}
      G_{IO}(s_l) \!= \!\frac{G_{LG}(s_l)}{1 + G_{LG}(s_l)}, \;\; \mbox{where}        \;\;
      G_{LG}(s_l) \!= \!\frac{G_h(s_l) e ^{-t_0 s_l}}{ms_l + b}
 \end{array}
\end{equation}
is the loop-gain function of the system. The qualitative Nyquist diagram of $G_{LG}(s_l)$ is shown in Fig.~\ref{nyquist_fig}. 
 The Nyquist criterion states that the closed-loop system is asymptotically stable as long as the Nyquist contour of $G_{LG}(s_l)$ does not touch or encircle the critical point $-1+j0$ \cite{Ogata}. Based on this, Fig.~\ref{nyquist_fig} clearly shows that a large  delay $t_0$ can lead to an unstable system.
The stability margin $M_{\varphi}$~\cite{Ogata} of the closed-loop human-robot interaction framework is shown in Fig.~\ref{PhaseMargin_bis}. 
%
Fig.~\ref{PM00_bis} shows that $M_{\varphi}>0$ is verified $\forall m,b$ when $t_0=0$, proving the system stability as predicted in the passivity proof of Sec.~\ref{Passivity_Analysis_section}. As $t_0$ increases, Fig.~\ref{PM01_bis} and Fig.~\ref{PM02_bis} show that the system becomes unstable for small values of $m$ and $b$ because $M_\varphi<0$.
This well agrees with the oscillations affecting the phase variable $s(t)$ in the experimental results of Fig.~\ref{epp} and Fig.~\ref{compare_sp}. 
It can be concluded that the parameters $m$ and $b$ in the phase calculation equation \eqref{F_h_eq} play an important role in determining the effort perceived by the user when moving along the desired trajectory, and they
need to be sufficiently large in order ensure the stability of the considered human-in-the-loop system.

}


{\black
\subsection{Insertion Task}
\label{Insertion_task}
}

\begin{figure*}[t]
    \centering
    \begin{subfigure}{0.19\linewidth}
        \includegraphics[width=\columnwidth]{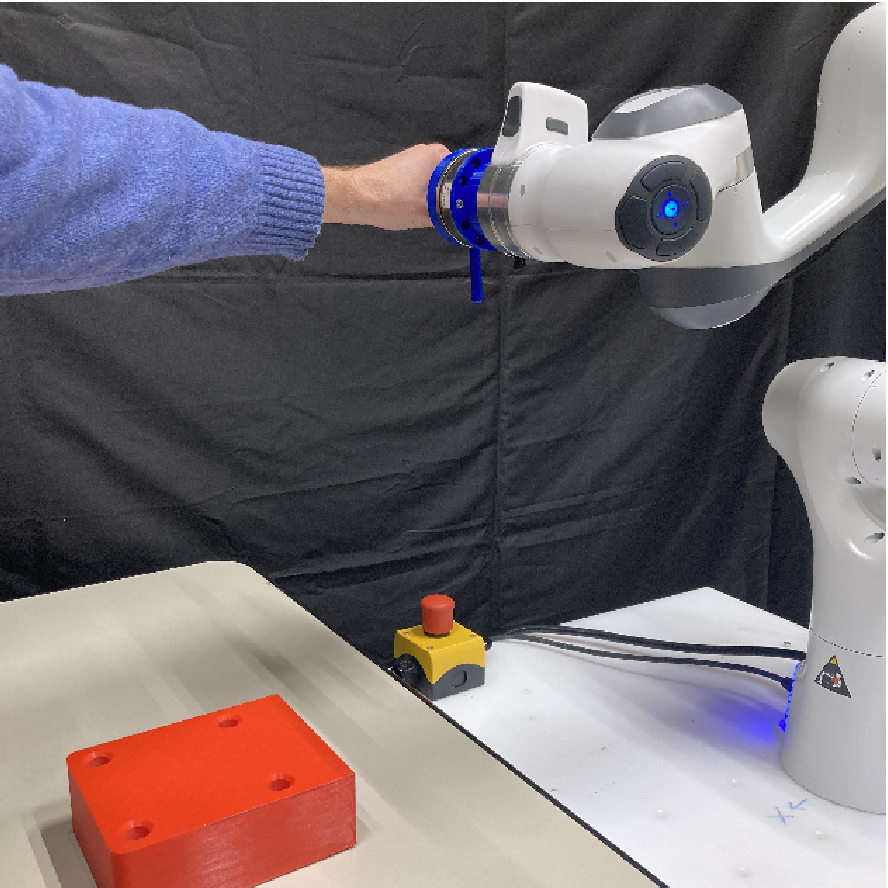}
        \caption{}
        \label{subfig:seq3_1}
    \end{subfigure}
    \begin{subfigure}{0.19\linewidth}
        \includegraphics[width=\columnwidth]{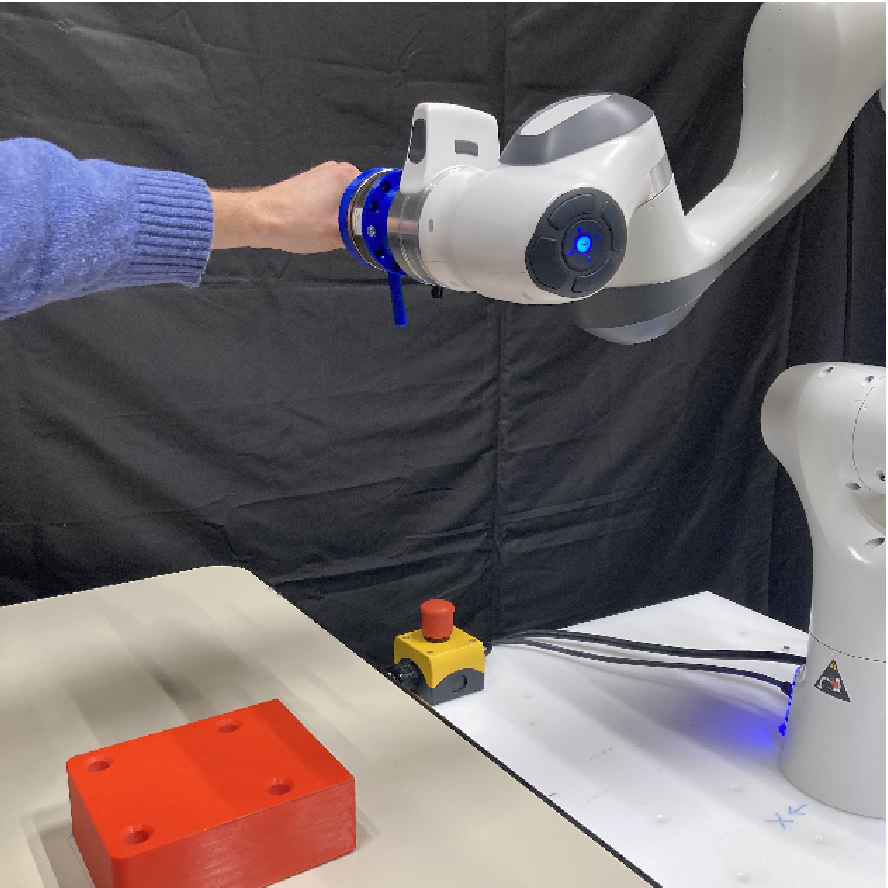}
        \caption{}
        \label{subfig:seq3_2}
    \end{subfigure}
    \begin{subfigure}{0.19\linewidth}
        \includegraphics[width=\columnwidth]{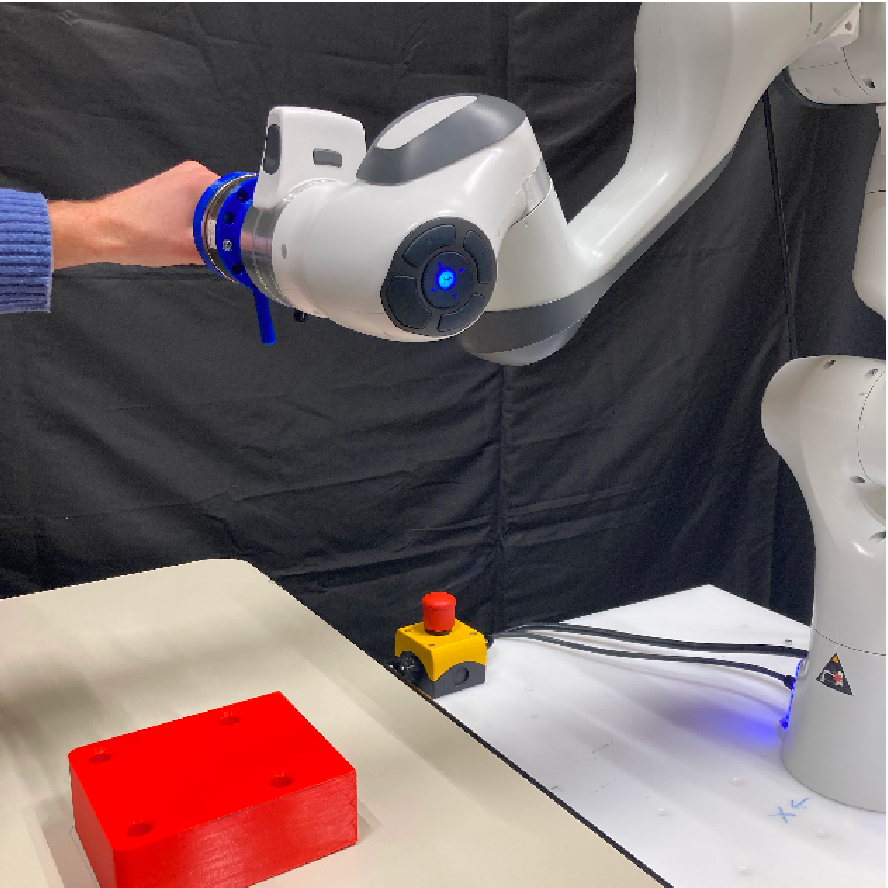}
        \caption{}
        \label{subfig:seq3_3}
    \end{subfigure}
    \begin{subfigure}{0.19\linewidth}
        \includegraphics[width=\columnwidth]{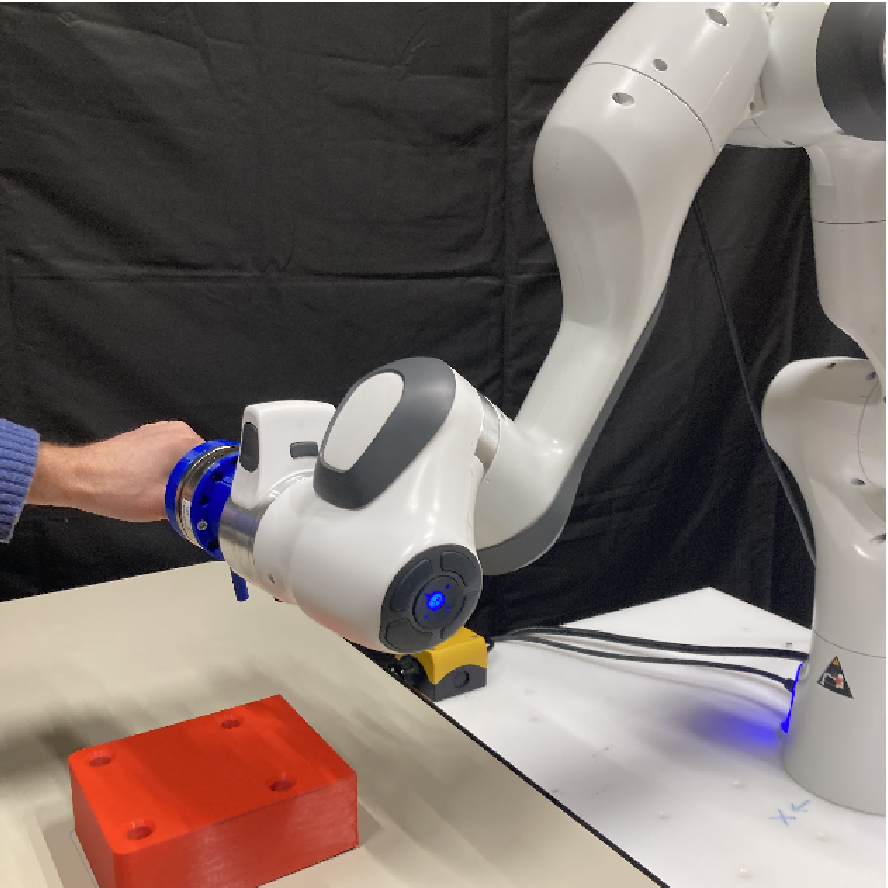}
        \caption{}
        \label{subfig:seq3_4}
    \end{subfigure}
    \begin{subfigure}{0.19\linewidth}
        \includegraphics[width=\columnwidth]{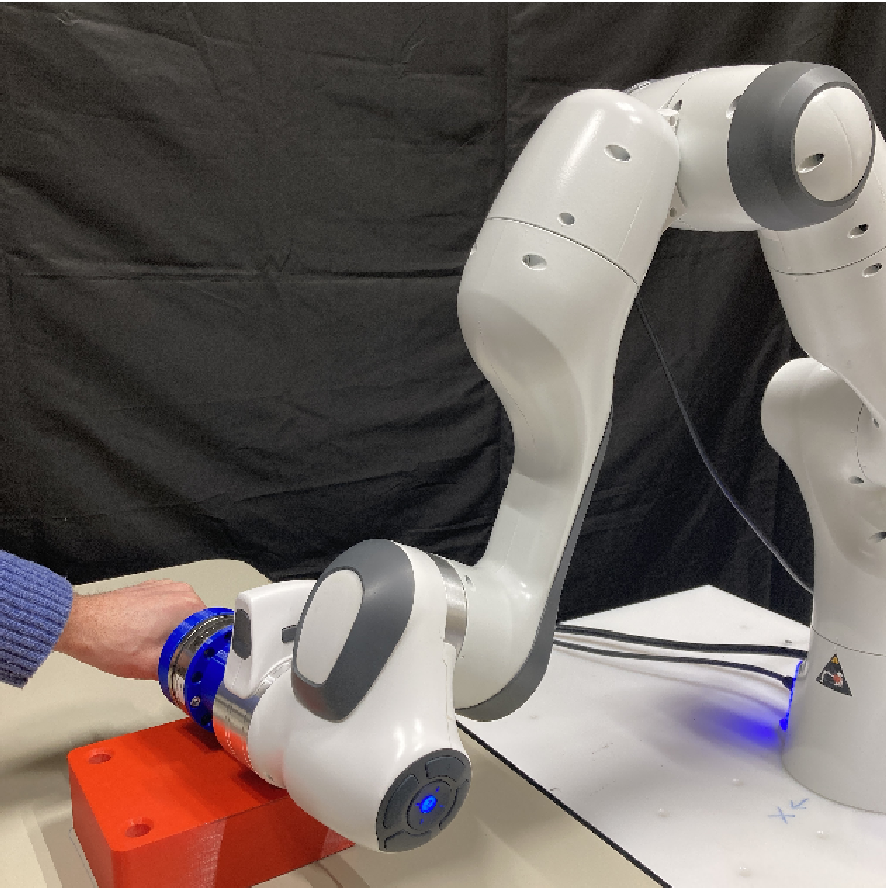}
        \caption{}
        \label{subfig:seq3_5}
    \end{subfigure}
    \caption{{\black Snapshots of the insertion task. During the demonstration by kinesthetic teaching, the user moves horizontally the robot over the target hole from its start position (Fig.~\ref{subfig:seq3_1}-~\ref{subfig:seq3_2}), then the user stops for few seconds (Fig.~\ref{subfig:seq3_3}) and moves vertically the end-effector tip toward the hole in the red box to complete the insertion task (Fig.~\ref{subfig:seq3_4}-~\ref{subfig:seq3_5}).}}
    \label{fig:insertion_task_snapshots}
\end{figure*}
\begin{figure*}[h!]
\psfrag{x}[c][t][0.9]{$x$ [m] }
\psfrag{y}[c][t][0.9]{$y$ [m] }
\psfrag{z}[c][t][0.9]{$z$ [m] }
    \centering
%
\begin{minipage}{0.42\linewidth}
    \begin{subfigure}[t]{0.9\linewidth}
    \centering
        \includegraphics[width=\textwidth]{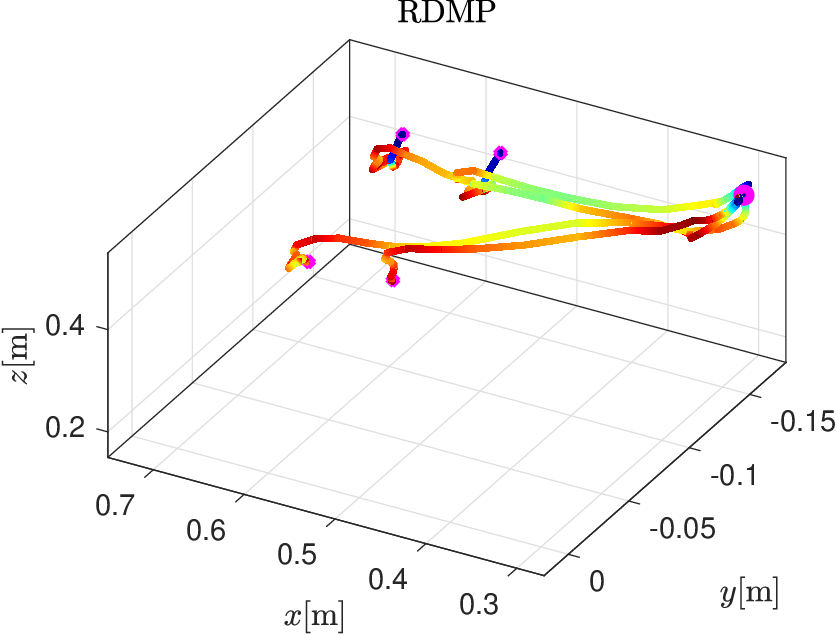}
        \caption{}
        \label{subfig:rdmp_3d}
    \end{subfigure}
    \begin{subfigure}[b]{0.9\linewidth}
    \centering
        \includegraphics[width=\textwidth]{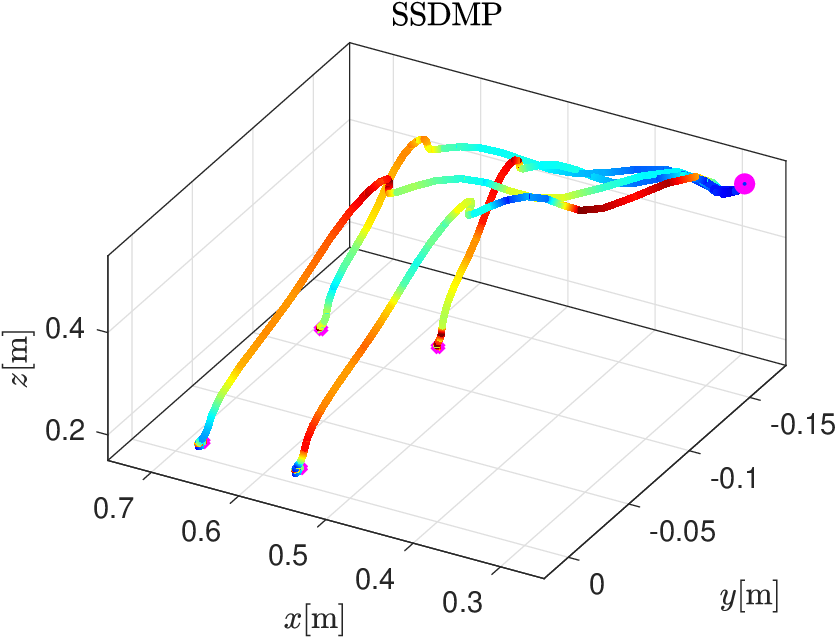}
        \caption{}
        \label{subfig:ssdmp_3d}
    \end{subfigure}
\end{minipage}
\begin{minipage}{0.42\linewidth}
    \begin{subfigure}[t]{0.9\linewidth}
    \centering
        \includegraphics[width=\textwidth]{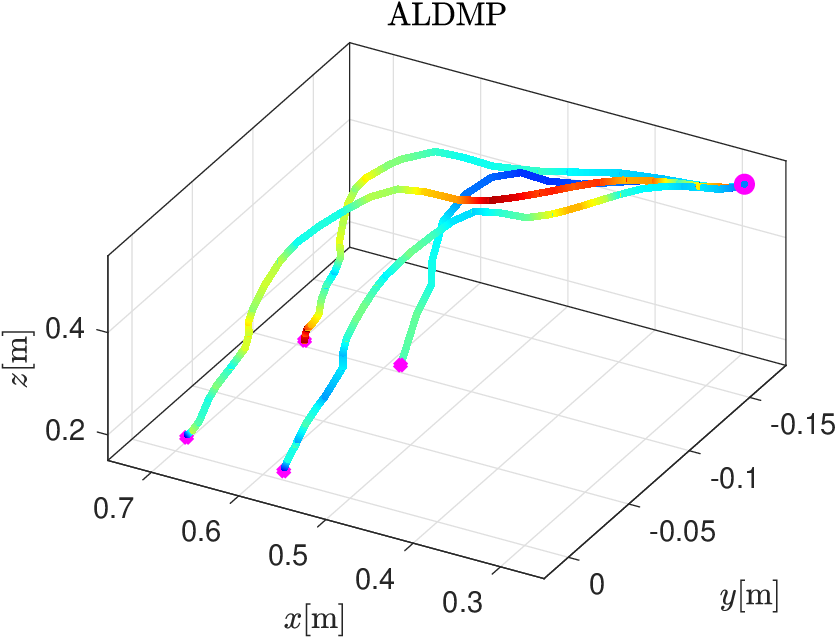}
        \caption{}
        \label{subfig:aldmp_3d}
    \end{subfigure}
    \begin{subfigure}[b]{0.9\linewidth}
    \centering
        \includegraphics[width=\textwidth]{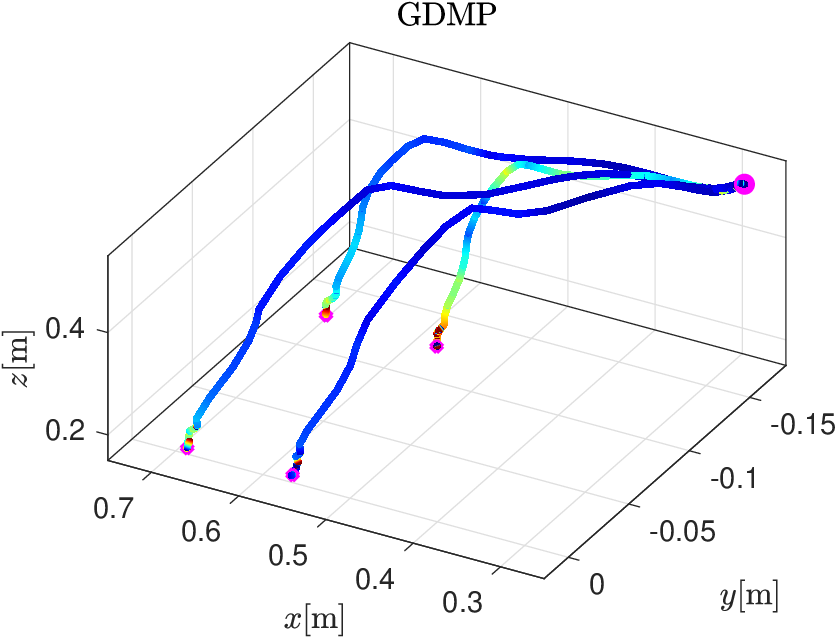}
        \caption{}
        \label{subfig:gdmp_3d}
    \end{subfigure}
\end{minipage}
\begin{minipage}{0.1\linewidth}
    \begin{subfigure}[t]{0.4\linewidth}
    \centering
        \includegraphics[width=\columnwidth]{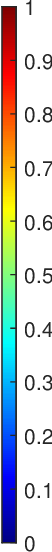}
        \caption{}
        \label{subfig:colorbar}
    \end{subfigure}
\end{minipage}
\caption{{\black Plots of the recorded trajectories when varying the goal position. Each position trajectory has been reported with varying color, which encodes the measured force norm $||\F_h||$ normalized to the maximum measured component inside the same experimental trial. } }
    \label{fig:insertion_task_3D_plots}
\end{figure*}
\begin{figure}[h]
\psfrag{aaaa}[][][0.8]{RDMP}
\psfrag{bbbb}[t][t][0.8]{SSDMP}
\psfrag{cccc}[][][0.8]{ALDMP}
\psfrag{dddd}[][][0.8]{GDMP}
\centering
\begin{subfigure}{0.9\linewidth}
\centering
    \includegraphics[width=\columnwidth]{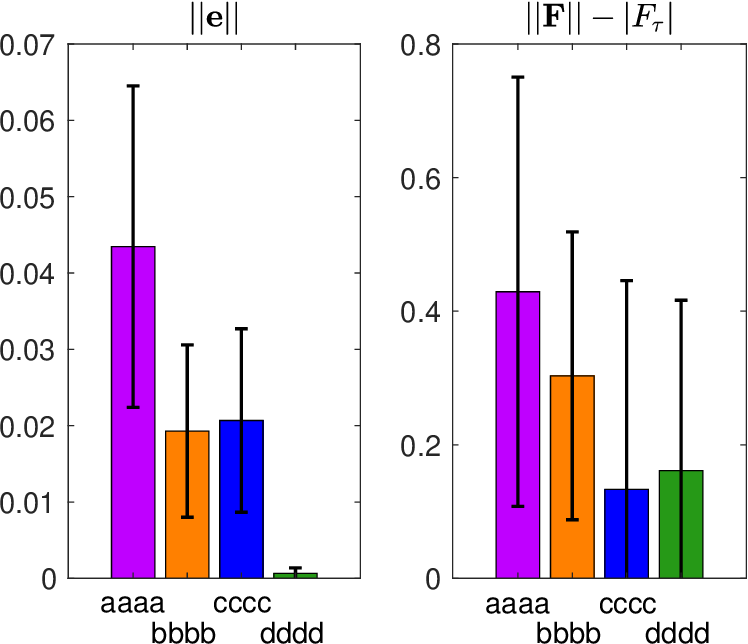}
    \caption{}
    \label{subfig:insertion_stat}
\end{subfigure}\\[2mm]
\begin{subfigure}{0.9\linewidth}
        \begin{center} 
    \begin{tabular}{@{}c@{\,}c@{\;\;}c@{\;\;}c@{}}
             &     \multirow{3}{*}{\small Reversibility} & {\small Insertion} & {\small Managing} \\
         &  & {\small Task} & {\small Pausing} \\
         &  & {\small Completion} & {\small Intervals} \\
    \hline
         RDMP & \gcheck & \rcross & \rcross \\
         SSDMP & \rcross & \gcheck & \gcheck \\
         ALDMP & \rcross & \gcheck & \gcheck\\
         GDMP & \gcheck & \gcheck & \gcheck\\        
     \hline
    \end{tabular}
    \end{center}
    \caption{}
    \label{subfig:table_insertion}
\end{subfigure} 
\caption{ {\black Comparison of GDMP with other DMP solutions in addressing the insertion task. (a) Obtained results in term of average norm of the error displacement $||\e||$, representing the spatial accuracy, and the average norm of the non-tangential force component $||\F_h||-|F_\tau|$,which reflects the undesired forces orthogonal to the task direction. (b) Final evaluation table. } }
\label{fig:Insertion_compare}
\end{figure}

{\black To further evaluate the GDMP architecture in Fig.~\ref{schematico}, a final co-manipulation task was conducted using $m=2$ and $b=17$ for robustness, as discussed previously. The task involved the guidance of the end-effector handle in order to insert its tip into one of the four holes in the red box shown in Fig.~\ref{fig:insertion_task_snapshots}. }

{\black The insertion task was initially demonstrated via kinesthetic guidance. Starting from the position in Fig.~\ref{subfig:seq3_1}, the robot was guided above the target hole (Fig.~\ref{subfig:seq3_3}), paused for few seconds, and then moved into the hole to complete the task (Fig.~\ref{subfig:seq3_4}-~\ref{subfig:seq3_5}). During the execution, the goal position of the DMP architecture was varied in order to select each of the four holes of the box.
The GDMP architecture was compared with three other DMP methodologies: reversible DMP (RDMP)~\cite{zoe1}, speed-scaled DMP (SSDMP)~\cite{nemec2018human}, and arc-length DMP (ALDMP)~\cite{GASPAR2018225}. Each approach was tested three times for all four holes of the box, and the results are presented in Fig.~\ref{fig:insertion_task_3D_plots}-~\ref{fig:Insertion_compare}. }

{\black Figure~\ref{fig:insertion_task_3D_plots} shows the 3D plots of the task execution for each DMP architecture. The position trajectories are color-coded based on the magnitude of the measured force norm $||\F_h||$ during the execution, where the force data have been acquired using a Schunk FT-AXIA force/torque sensor. To allow the comparison across the different DMP methods, all executions were normalized to their maximum measured force. A colorbar for the force magnitude is provided in Fig.~\ref{subfig:colorbar}. }

{\black For what concerns the RDMP case shown in Fig.~\ref{subfig:rdmp_3d}, a high force demand is evident due to the lack of arc-length parametrization, which prevents proper tangential projection of the forces along the trajectory. This issue becomes critical near the stopping interval recorded during the demonstration, where the tangential component is undefined, making it impossible for the user to complete the insertion task. 
In the SSDMP case shown in Fig.~\ref{subfig:ssdmp_3d}), although a higher force demand is observed, the user successfully completes the task. This is because the SSDMP autonomously progresses its phase, actively guiding the user. However, the tests have shown that the SSDMP's phase velocity adjustment based on the user's feedback lacks reactivity, leading to difficulty switching between active and passive guidance, and resulting in a higher force demands during the execution. }

{\black Significantly lower force measurements are observed for the ALDMP case shown in Fig.~\ref{subfig:aldmp_3d}. This is because the arc-length parametrization used by ALDMP allows a consistent computation of the tangential component of the desired curve at each instant, ensuring the correct projection of the measured force $\F_h$. However, the color code shows frequent force variations, as the ALDMP algorithm does not maintain a constant tangential norm, unlike the proposed SS algorithm.
In the GDMP case shown in Fig.~\ref{subfig:gdmp_3d}, the SS algorithm ensures that the tangential component of the geometric curve has unitary norm, as guaranteed by the filtering action obtained in property~\eqref{unitary_norm} which ensures the regularity of the curve. This results in consistent projection of $\F_h$ throughout the task, reducing the user's effort when moving the robot, as evident from the trajectory color code.}

{\black The task execution is further analyzed in the results of Fig.~\ref{fig:Insertion_compare}. The statistical results are reported in Fig.~\ref{subfig:insertion_stat} based on: (i) the norm of the displacement error $||\e||$, where $\e(t) = \y(t)-\y_m(t)$; and (ii) the difference $||\F_h||-|F_\tau|$, representing the force used to keep the end-effector near the trajectory $\y$ provided by the DMP solution.
The proposed GDMP achieves more accurate position tracking, as shown in Fig.~\ref{fig:insertion_task_3D_plots}, thanks to its more effortless guidance compared to other DMP methods. Additionally, the lower difference $||\F_h||-|F_\tau|$ indicates that, like ALDMP, the GDMP provides more intuitive guidance. This is because a greater portion of the applied force is projected tangentially to the curve, making the robot guidance easier to follow. }

{\black The limitations of the analyzed DMP approaches when compared with GDMP are summarized in the table in Fig.~\ref{subfig:table_insertion}, where ``Reversibility'', ``Insertion Task Completion'', and ``Managing Pausing Intervals'' refer to the DMP's ability to be reversible, complete the insertion task, and handle pausing intervals recorded during the demonstration, respectively. 
The RDMP fails to accomplish the task due to its reliance on time parametrization, which prevents it from managing pausing intervals. The SSDMP and ALDMP lack reversibility, making them unsuitable for co-manipulation tasks. While the SSDMP can handle pausing intervals by actively guiding the user and resuming movement after the pause, its execution remains time-dependent on the demonstration, posing limitations. 
In contrast, the proposed GDMP satisfies all the requirements, demonstrating its enhanced performance over the other approaches. }

\section{Conclusions}\label{Conclusions_sect}

The Canonical System of the DMP controls the evolution of phase variable, and thus of the forcing term, shaping the attractor landscape for the desired output dynamics. While standard DMP ensure time modulation, no unified framework exists in the literature to compactly handle multiple phase profiles without significantly altering the DMP formulation. The study proposed in this paper introduces a new concept called Geometric Dynamic Movement Primitives (GDMP), thanks to which the phase variable can be freely chosen depending on the desired application. The concept of GDMP is based on the proposed spatial sampling algorithm, which decouples the demonstrated curve from its timing law thus allowing to generate an arc-length parameterized geometric path. {\black The proposed spatial sampling algorithm guarantees the parameterized curves to always be regular, thus ensuring a consistent projection of the human force throughout the task in a
human-in-the-loop scenario.}
{\black The effectiveness of the GDMP has been demonstrated through two main applications. The first one is an offline optimization problem for minimum task duration, subject to velocity and acceleration constraints, which has highlighted the GDMP's superiority over previous DMP solutions thanks to the separation of path and velocity. The second application of GDMP is human-in-the-loop involving different co-manipulation tasks. For this application, 
an analytical passivity analysis and an analytical/experimental stability analysis have been carried out. 
Finally, the proposed human-in-the-loop architecture is further validated with reference to an insertion task, where GDMP is experimentally compared against other DMP solutions and is shown to consistently outperform them.
}

\setcounter{secnumdepth}{0}
\section{Funding}
The work was partly supported by the University of Modena and Reggio Emilia
through the action FARD (Finanziamento Ateneo Ricerca Dipartimentale) 2022/2023 and 2023/2024, and funded under the National Recovery and Resilience Plan (NRRP), Mission 04 Component 2 Investment 1.5 – NextGenerationEU, Call for tender n. 3277 dated 30/12/2021
Award Number:  0001052 dated 23/06/2022.

\renewcommand{\theequation}{A.\arabic{equation}}
\setcounter{equation}{0} \setcounter{secnumdepth}{0}
\section{Nomenclature}\label{Nomenc_sect}

\begin{supertabular}{@{\,}c@{\hspace{3.5mm}}l@{\,}}
\rowcolor{Gray}
$ \g  $  & goal position \\
$ \y_r(t),\dy_r(t),\ddy_r(t) $  & demonstrated trajectories \\ \rowcolor{Gray}
$ T_f $  & task duration \\
$ T $ & sampling-time \\\rowcolor{Gray}
$ \f(s(t)) $ & forcing term \\
$ \f^\star(s(t)) $ & parameterized forcing term \\\rowcolor{Gray}
$ s(t) $ & phase variable \\
$ \yp(t), \dyp(t), \ddyp(t)$ & parameterized task trajectories \\\rowcolor{Gray}
$ \y(t), \dy (t), \ddy(t) $ & GDMP trajectories  \\ 
$ \alpha / \A $ & friction coefficient / matrix \\\rowcolor{Gray}
$ \alpha \beta / \A \B $ & spring stiffness / matrix \\
$ \eta / \E $ & scaling coefficient / matrix \\\rowcolor{Gray}
$ \qp(t), \dqp(t), \ddqp(t) $ & 
parameterized joint trajectories 
\\\rowcolor{Gray}
$ \y_m(t), \dy_m(t), \ddy_m(t) $ & actual robot trajectories \\ 
$ \Y_T $ & input samples matrix \\\rowcolor{Gray}
$ \t_T $ & time vector associated with $ \Y_T $ \\
$ \y_L(t) $ & linearly interpolated trajectory \\\rowcolor{Gray}
$ \Delta $ & spatial sampling interval \\
$ \yfk( \sfk ) $ & spatially-sampled trajectory \\\rowcolor{Gray}
$ \sfk $ & phase vector \\
$ \t_\Delta $ & time vector associated with $\sfk$ \\\rowcolor{Gray}
$\Yfk$ & output samples matrix \\
$ \Delta M $ & length of $\y_r(t)$ \\\rowcolor{Gray}
$ m $ & mass parameter \\
$ b $  & damping parameter \\\rowcolor{Gray}
$ \F_h(t) $&  applied user force vector \\
$ F_{\tau}(t) $ & user tangential force \\\rowcolor{Gray}
$ s_l $ & Laplace complex variable \\
\end{supertabular}

\bibliographystyle{elsarticle-num}
\bibliography{references}

\end{document}